\documentclass[twoside]{article}

\usepackage[accepted]{aistats2025}

\usepackage{graphicx} 

\usepackage[utf8]{inputenc} 
\usepackage[T1]{fontenc}    
\usepackage[colorlinks,linktoc=all]{hyperref}       
\usepackage[dvipsnames]{xcolor}
\hypersetup{linkcolor=MidnightBlue,citecolor=MidnightBlue,urlcolor=MidnightBlue}
\usepackage{url}            
\usepackage{booktabs}       
\usepackage{amsfonts}       
\usepackage{nicefrac}       
\usepackage{microtype}      
\usepackage{xcolor}         

\usepackage{natbib}

\usepackage{algorithmic,algorithm}
\usepackage{nicefrac}

\usepackage{amsmath, amssymb, amsthm, bbm, dsfont, mathrsfs}
\usepackage{subfiles, subcaption}
\usepackage{color, verbatim, graphicx}
\usepackage{enumitem}
\usepackage{mathtools}
\usepackage[nameinlink]{cleveref}
\creflabelformat{equation}{#1#2#3}
\crefname{equation}{eq.}{eqs.}
\Crefname{equation}{Eq.}{Eqs.}
\usepackage{multirow,multicol}

\usepackage{xr}

\makeatletter
\newcommand*{\addFileDependency}[1]{
  \typeout{(#1)}
  \@addtofilelist{#1}
  \IfFileExists{#1}{}{\typeout{No file #1.}}
}
\makeatother

\numberwithin{equation}{section}
\numberwithin{figure}{section}
\numberwithin{table}{section}

\theoremstyle{definition}
\newcommand{\SigmaFull}{\Sigma_{t+\scriptscriptstyle\frac{1}{2}}}

\renewcommand{\epsilon}{\varepsilon}

\newcommand{\given}{\,|\,}

\newcommand{\norm}[1]{\left\lVert#1\right\rVert}
\newcommand{\trace}{\text{tr}}

\newcommand{\A}{\mathcal{A}}

\newcommand{\C}{\mathcal{C}}
\newcommand{\D}{\mathscr{D}}

\renewcommand{\L}{\mathcal{L}}
\newcommand{\N}{\mathcal{N}}
\renewcommand{\O}{\mathcal{O}}
\newcommand{\Q}{\mathcal{Q}}

\renewcommand{\SS}{\mathbb{S}}

\newcommand{\ZZ}{\reals^D}

\newcommand{\E}{\mathbb{E}}

\newcommand{\Cov}{\text{Cov}}
\newcommand{\KL}{\text{KL}}

\newcommand{\reals}{\mathbb{R}}

\newtheorem*{theorem*}{Theorem}
\newtheorem*{proposition*}{Proposition}
\newtheorem*{lemma*}{Lemma}
\newtheorem*{corollary*}{Corollary}

\usepackage{thmtools}

\definecolor{shade}{rgb}{0.9,0.9,0.9}

\declaretheoremstyle[
spaceabove=6pt, spacebelow=6pt,
postheadspace=1em,
headfont=\normalfont\bfseries,
notefont=\mdseries, notebraces={(}{)},
bodyfont=\normalfont,
]{mystyle}

\declaretheoremstyle[
    spaceabove=-6pt,  spacebelow=6pt,
    headfont=\normalfont\bfseries,
    bodyfont = \normalfont,
    postheadspace=1em,
    qed=$\blacksquare$,
    headpunct={:}]{myproofstyle} 

\DeclareUnicodeCharacter{03BC}{$\mu$}
\DeclareUnicodeCharacter{03A3}{$\Sigma$}
\DeclareUnicodeCharacter{0393}{$\Gamma$}
\DeclareUnicodeCharacter{03BB}{$\lambda$}

\usepackage{eqparbox}

\begin{document}


\twocolumn[

\runningtitle{Batch, match, and patch: low-rank approximations for score-based variational inference}

\aistatstitle{Batch, match, and patch:\
low-rank approximations \\
for score-based variational inference}

\aistatsauthor{Chirag Modi$^*$
\And Diana Cai$^*$ 
\And  Lawrence K.\ Saul }

\aistatsaddress{
Center for Computational\\Mathematics, Flatiron Institute\\
Center for Cosmology and Particle\\Physics, New York University\\
\And
Center for Computational\\Mathematics, Flatiron Institute
\And Center for Computational\\Mathematics,  Flatiron Institute} ]

\begin{abstract}
Black-box variational inference (BBVI) scales poorly to high-dimensional problems when it is used to estimate a multivariate Gaussian approximation with a full covariance matrix. In this paper, we extend the \emph{batch-and-match} (BaM) framework for score-based BBVI to problems where it is prohibitively expensive to store such covariance matrices, let alone to estimate them. Unlike classical algorithms for BBVI, which use stochastic gradient descent to minimize the reverse Kullback-Leibler divergence, BaM uses more specialized updates to match the scores of the target density and its Gaussian approximation. We extend the updates for BaM by integrating them with a more compact parameterization of full covariance matrices. In particular, borrowing ideas from factor analysis, we add an extra step to each iteration of BaM---a \emph{patch}---that projects each newly updated covariance matrix into a more efficiently parameterized family of diagonal plus low rank matrices. We evaluate this approach on a variety of synthetic target distributions and real-world problems in high-dimensional inference.
\end{abstract}

\section{Introduction}

Many statistical analyses can be formulated as problems in Bayesian inference. In this formulation, the goal is to compute the posterior distribution over latent random variables from observed ones. But difficulties arise when the latent space is of very high dimensionality,~$D$. In this case, even if the underlying models are simple---e.g., Gaussian---it can be prohibitively expensive to store let alone estimate a $D\times D$ covariance matrix.

These difficulties are further compounded when the underlying models are not Gaussian. In this case, though, a multivariate Gaussian can sometimes be used to approximate a posterior distribution that is otherwise intractable. This idea is the basis of variational inference (VI) \citep{jordan1999vi,wainwright2008graphical,blei2017vi}, which in this setting seeks the Gaussian approximation with the lowest divergence to the posterior. There are well-developed software packages to compute these approximations that make very few assumptions about the form of the posterior distribution; hence  this approach is known as black-box variational inference (BBVI) \citep{ranganath2014black,kingma2013auto,titsias2014doubly,kucukelbir2017automatic}.
BBVI, however, also scales poorly to high-dimensional problems when it is used to estimate a multivariate Gaussian approximation with a full covariance matrix \citep{ko2024provably}.

One hopeful approach is to parameterize a large covariance matrix as the sum of a diagonal matrix and a low-rank matrix. This idea is used in maximum likelihood models for factor analysis, where the diagonal and low-rank matrices can be estimated by an Expectation-Maximization (EM) algorithm~\citep{rubin1982algorithms,ghahramani1996algorithm,saul2000maximum}. This parameterization can also be incorporated into BBVI algorithms that use stochastic gradient methods to minimize the reverse Kullback-Leibler (KL) divergence; in this case, the required gradients can be computed automatically by the chain rule~\citep{miller2017variational,ong2018gaussian}. This approach overcomes the poor scaling with respect to the dimensionality of the latent space, but it still relies on the heuristics and hyperparameters of stochastic gradient methods. In particular, the optimization itself does not leverage in any special way the structure of factored covariance matrices.

In this paper, we extend the \emph{batch-and-match} (BaM) framework for score-based BBVI \citep{cai2024} to these problems. Unlike classical algorithms for BBVI, BaM does not use gradient descent to minimize a KL divergence; it is based on more specialized updates to match the scores of the target density and its Gaussian approximation. We augment these original updates for BaM with an additional step---a \emph{patch}---that projects each newly updated covariance matrix into a more efficiently parameterized family of diagonal plus low rank matrices. This patch is closely based on the EM updates for maximum likelihood factor analysis, and the resulting algorithm scales much better to problems of high dimensionality.
We refer to this variant of the algorithm as \textit{patched batch-and-match} (pBaM)\footnote{A Python implementation of pBaM is available at \href{https://github.com/modichirag/GSM-VI/}{https://github.com/modichirag/GSM-VI/}.}.

The organization of this paper is as follows. In \Cref{sec:background}, we review the basic ideas of BBVI and BaM. In \Cref{sec:patch}, we derive the patch for BaM with structured covariance matrices. In \Cref{sec:related}, we summarize related work on these ideas. In \Cref{sec:experiments}, we evaluate the pBaM algorithm on a variety of synthetic target distributions and real-world problems in high-dimensional inference. Finally, in \Cref{sec:conclusion}, we conclude and discuss directions for future work.

%
\section{Score-based variational inference}
\label{sec:background}

In this section, we provide a brief review of BBVI, score-based VI, and the BaM algorithm for score-based BBVI with multivariate Gaussian approximations.

\subsection{Black-box variational inference}

The goal of VI is to approximate an intractable target density $p$ by its closest match $q$
in a variational family~$\Q$.
The goal of BBVI is to compute this approximation in a general way that makes very few assumptions about the target $p$. Typically, BBVI assumes only that it is possible to evaluate the \emph{score} of $p$ (i.e., $\nabla_z \log p(z)$) at any point $z$ in the latent space.

As originally formulated, BBVI attempts to find the best approximation $q\!\in\!\mathcal{Q}$ by minimizing the reverse KL divergence, $\text{KL}(q||p)$, or equivalently by
maximizing the evidence lower bound (ELBO).
One popular approach for ELBO maximization is the automatic differentiation framework for variational inference (ADVI),
which is based on stochastic gradient-based optimization.
ADVI \citep{kucukelbir2017automatic}, which uses the reparameterization trick to cast the gradient of the ELBO as an 
expectation,
constructs a stochastic estimate of this gradient.
The main strength of this approach is its generality:
it can be used with any variational family that lends itself to reparameterization. But this generality also comes at a cost: the approach may converge very slowly for richer families of variational approximations---including, in particular, the family of multivariate Gaussian densities with full (dense) covariance matrices.

\subsection{Score-based approaches to BBVI}

Recently, several algorithms have been proposed for BBVI that aim to match the scores of the target density and its variational approximation~\citep{yang2019variational,zhang2018variational,Modi2023,cai2024,cai2024b}. 
Some of these score-based approaches are specially tailored to Gaussian variational families~\citep{Modi2023,cai2024}, and they exploit the particular structure of Gaussian distributions to derive more specialized updates for score-matching.

These approaches find the best Gaussian approximation by minimizing
a weighted score-based divergence:
\begin{align}
\label{eq:scorediv}
    \D(q; p) \!:=\! \int\! \norm{\nabla\log q(z) \!-\! \nabla\log p(z)}^2_{\Cov(q)} \, q(z) \, dz,
\end{align}
where $\norm{x}_\Sigma := \sqrt{x^\top \Sigma x}$ and $q$ is assumed to belong to the variational family $\Q := \{\N(\mu,\Sigma)\!:\! \mu\!\in\!\reals^D,\Sigma\!\in\! \SS_{++}^D\}$ of multivariate Gaussian distributions.

One of these approaches is the batch-and-match (BaM) algorithm for score-based VI. BaM uses a KL-regularized proximal point update to optimize the above divergence.
BaM  avoids certain heuristics of gradient-based optimization because its proximal point update can be solved in closed form.
The BaM updates reduce as a special case to an earlier algorithm for Gaussian score matching \citep{Modi2023}; the latter is equivalent to BaM with a batch size of one and an infinite learning rate.
Notably, both of these approaches have been empirically observed
to converge faster than ADVI with a full covariance. The ``patch'' step in this paper can be applied both to Gaussian score matching and BaM, but we focus on the latter because it contains the former as a special case.

\subsection{Batch-and-match variational inference}

We now review the BaM framework for BBVI in more detail. BaM iteratively updates its variational parameters
by minimizing an objective with two terms: one is a score-based divergence between $q$ and $p$, estimated from a batch of $B$ samples, while the other is a regularizer that penalizes overly aggressive updates. Specifically, the update is given by
\begin{align}
\label{eq:batch}
    q_{t+1} = \arg\min_{q \in \Q} \widehat\D_{q_t}(q; p) + \tfrac{2}{\lambda_t}
    \KL(q_t; q),
\end{align}
where $\widehat\D_{q_t}(q; p)$ is a stochastic estimate
of the score-based divergence in \Cref{eq:scorediv}
based on $B$ samples $\{z_b \sim q_t\}_{b=1}^B$. We compute this stochastic estimate~as
\begin{align}
\label{eq:empiricalscore}
\widehat\D_{q_t}(q; p) :=  \sum_{b=1}^B \big\|\nabla\log q(z_b) \!-\! \nabla\log p(z_b)\big\|^2_{\Cov(q)}.
\end{align}
Note also that the prefactor of the KL regularizer in \Cref{eq:batch} contains a \emph{learning rate} parameter $\lambda_t > 0$.

The optimization in \Cref{eq:batch} has a closed-form solution, and this solution forms the basis for BaM.
Each iteration of BaM alternates between a \emph{batch step} and a \emph{match step}. The \emph{batch step} collects $B$ samples $\{z_b\sim q_t\}_{b=1}^B$, evaluates their scores~$g_b := \nabla \log p(z_b)$, and then computes the following statistics:
\begin{align}
\label{eq:statistics}
    \overline{z}_{} &= \tfrac{1}{B} \sum_{b=1}^B z_b,
                \qquad
                C_{} = \tfrac{1}{B} \sum_{b=1}^B (z_b-\overline{z}) (z_b-\overline{z})^\top
                \\
        \label{eq:statistics2}
                \overline{g}_{} &= \tfrac{1}{B} \sum_{b=1}^B g_b, \qquad
                \Gamma_{} = \tfrac{1}{B} \sum_{b=1}^B (g_b-\overline{g}) (g_b-\overline{g})^\top.
\end{align}
Finally, the batch step uses these statistics to compute the empirical score-based divergence
$\widehat\D_{q_t}(q; p)$ in \Cref{eq:empiricalscore}.
{We note for future reference that the matrices $C$ and~$\Gamma$ in these statistics are constructed as the sum of $B$ \mbox{rank-1} matrices; thus \textit{the ranks of $C$ and $\Gamma$ are less than or equal to the batch size~$B$}.}
 
The \emph{match step} of BAM computes updated variational parameters $\Sigma_{t+1}$ and $\mu_{t+1}$ that minimize the score-matching objective in \Cref{eq:empiricalscore}.
In what follows we present the updates for $B\!<\!D$, which is the relevant regime for high-dimensional problems. 
We begin by introducing two intermediate matrices $U$ and $V$, defined~as
       \begin{align}
            \label{eq:U}
            U &:=  \lambda_t \Gamma + \tfrac{\lambda_t}{1+\lambda_t} \, \overline{g}\, \overline{g}^\top
            \\
            \label{eq:V}
            V &:=  \Sigma_t + \lambda_t C + \tfrac{\lambda_t}{1+\lambda_t}
            (\mu_t-\overline{z})(\mu_t-\overline{z})^\top.
        \end{align}
In fact, it is not necessary to store the dense $D\times D$ matrix $U$; it suffices to work with any \text{tall and thin} matrix $Q\!\in\!\reals^{D\times(B+1)}$ satisfying
\begin{equation}
U=QQ^\top.
\label{eq:UQQt}
\end{equation}
One solution for $Q$ satisfying the above can be immediately deduced from the expressions for $\Gamma$ and $U$ in \Cref{eq:statistics2,eq:U}; in particular, it is obtained by concatenating the mean and centered scores of the batch:
\begin{equation}
\label{eq:Q}
Q = \left[\sqrt{\tfrac{1}{B}}(g_1\!-\!\overline{g}),\ \ldots, \sqrt{\tfrac{1}{B}}(g_B\!-\!\overline{g}),\
\sqrt{\tfrac{\lambda_t}{1+\lambda_t}}\overline{g} \right].
\end{equation}
Finally, in terms of these matrices, BaM updates the covariance as
\begin{equation}
  \label{eq:sigmaupdate}
   \Sigma_{t+1} \!=\! V \!-\! V^\top Q \big[\tfrac{1}{2} I \!+\! \left(Q^\top V Q + \tfrac{1}{4} I\right)^\frac{1}{2}\big]^{-2} \! Q^\top V,
\end{equation}
where the thinness of $Q$ helps to perform this update more efficiently. After this covariance update, BaM updates the mean as
\begin{align}
{\mu}_{t+1} =
            \tfrac{1}{1+\lambda_t} \mu_t + \tfrac{\lambda_t}{1+\lambda_t}\left(\Sigma_{t+1} \, \overline{g} +
            \overline{z}\right),
\end{align}
where $\Sigma_{t+1}$ is the update in \Cref{eq:sigmaupdate}. The new parameters $\Sigma_{t+1}$ and $\mu_{t+1}$ from these updates in turn define the new variational approximation $q_{t+1}$.

{BaM differs from previous approaches to BBVI that optimize a KL-based objective using gradient descent. Gradient-based approaches rely at each iteration on an implicit linearization of the objective function that they are trying to optimize. By contrast, each BaM iteration constructs a proximal-point objective function whose global optimum can be computed in closed form and without resorting to an implicit linearization.}

Finally, we note that the BaM update in  \Cref{eq:sigmaupdate}
takes $\O(D^2 B + B^3)$ computation for dense covariance matrices $\Sigma_t$.
Since this update scales quadratically in~$D$, it may be too expensive to compute for very high dimensional problems. (In fact, it may even be too expensive to store the covariance matrix $\Sigma_t$.) In the next section, we introduce a \textit{patch} at each iteration of BaM to avoid this quadratic scaling.


\begin{algorithm}[t]
\begin{algorithmic}[1]
    \small
    \STATE{\textbf{Input:} Iterations $T$, batch size $B$,
    inverse regularization \mbox{$\lambda_t\!>\!0$},
    target score function $s: \ZZ \rightarrow \reals^D$,
        initial variational mean $\mu_0\!\in\! \reals^D$ and
    covariance matrix $\Sigma_0 = \Psi_0 + \Lambda_0\Lambda_0^\top$}.
    \FOR{$t=0,\ldots,T\!-\!1$}
    \STATE Sample batch  $z_b \sim \mathcal{N}(\mu_t,\Psi_t\!+\!\Lambda_t\Lambda_t)$ for $b\!=\! 1,\ldots, B$
\STATE Evaluate scores $g_{b} = s(z_b)$ for $b = 1,\ldots, B$
        \STATE Compute the batch mean statistics $\overline{z}, \overline{g}\! \in\! \reals^D$ given~by 
        \vspace{-8pt}
        $$\overline{z}= \tfrac{1}{B} \sum_{b=1}^B z_b\quad\mbox{and}\quad\overline{g} = \tfrac{1}{B} \sum_{b=1}^B g_b.$$
         \vspace{-4pt}
         \STATE Compute the intermediate matrices 
         \vspace{-5pt}
        \begin{align*}
            Q\! &=\! \left[\sqrt{\tfrac{1}{B}}(g_1\!-\!\overline{g}),\ \ldots, \sqrt{\tfrac{1}{B}}(g_B\!-\!\overline{g}),\ 
            \sqrt{\tfrac{\lambda_t}{1+\lambda_t}}\overline{g}\right], \\
            R\! &=\! \left[\sqrt{\tfrac{\lambda_t}{B}}(z_1\!-\!\bar{z}),...,\sqrt{\tfrac{\lambda_t}{B}}(z_B\!-\!\bar{z}),\sqrt{\tfrac{\lambda_t}{1+\lambda_t}}
    (\mu_t\!-\!\overline{z}), \Lambda_t\right],\\
            V\! &=\! \Psi_t + RR^\top.
        \end{align*}
        \vspace{-12pt}
    \STATE Update variational parameters (``match'' and ``patch'')
       \vspace{-15pt}
        \begin{align*}
            {\Sigma}_{t+\scriptscriptstyle\frac{1}{2}} 
            &=
            V \!-\! V^\top Q \bigg[\tfrac{1}{2} I \!+\! \big(Q^\top V Q + \tfrac{1}{4} I\big)^\frac{1}{2}\bigg]^{-2} \!\!\!\! Q^\top V,
            \\
            {\Sigma}_{t+1} 
            &=
            \texttt{Patch}\left[{\Sigma}_{t+\scriptscriptstyle\frac{1}{2}} \right]
            \\
            {\mu}_{t+1} &=
            \tfrac{1}{1+\lambda_t} \mu_t + \tfrac{\lambda_t}{1+\lambda_t}\left(\Sigma_{t+1} \, \overline{g} +
            \overline{z}\right)
        \end{align*}
        \vspace{-12pt}
     \ENDFOR
    \STATE \textbf{Output:}
    mean $\mu_T$ and covariance parameters $\Psi_T,\Lambda_T$
    \end{algorithmic}
    \caption{Batch, match, and patch BBVI
    }
    \label{alg:pBaM}
\end{algorithm}

\begin{algorithm}[t]
\caption{\texttt{Patch} step }
\label{alg:patch}
\begin{algorithmic}[1]
    \small
    \STATE{\textbf{Input:} 
    Dense covariance $\SigmaFull$ (represented implicitly by $\Psi_t,\,R$ and $Q$),
    initial 
    parameters $\Lambda_0, \Psi_0$, 
    momentum $\eta$, tolerance $\epsilon$, maximum number of steps $N$
    }
    \vspace{1pt}
    
    \STATE $\mathrm{KL}_0 = \log(|\Lambda_0\Lambda_0^\top + \Psi_0|) + \mathrm{tr}\big((\Lambda_0\Lambda_0^\top + \Psi_0)^{-1}\SigmaFull\big) $
   \FOR{$\tau=0,\ldots,N\!-\!1$}
   \vspace{2pt}
   \STATE 
   Compute intermediate matrix of size $K \times D$
   \vspace{-6pt}
    \[
    \beta_\tau =  \Lambda_\tau^\top
    \Psi_\tau^{-1} [I 
    \!-\! 
    \Lambda_\tau(I+\Lambda_\tau^\top \Psi_\tau^{-1} \Lambda_\tau)^{-1} \Lambda_\tau^\top \Psi_\tau^{-1}]
    \]
    \vspace{-13pt}
    \STATE Update low rank and diagonal parameters
    \vspace{-8pt}
    \begin{align*}
    \Lambda_{\tau+1} &=  \SigmaFull \beta_\tau^\top
    (\beta_\tau \SigmaFull \beta_\tau^\top + I - \beta_\tau\Lambda_\tau)^{-1}
\\
    \Psi_{\tau+1}
    &=
    \text{diag}((I -  \Lambda_{\tau+1} \beta_\tau) \SigmaFull)
    \end{align*}
    
\vspace{-8pt}
    \vspace{-8pt}
    \begin{align*}
    \Lambda_{\tau+1} &= (1-\eta) \Lambda_{\tau} + \eta \Lambda_{\tau+1}\mbox{\hspace{17ex}} \\ 
    \Psi_{\tau+1} &= (1-\eta) \Psi_{\tau} + \eta \Psi_{\tau+1}
    \end{align*}
    \vspace{-15pt}
    
    \STATE Compute 
    \vspace{-8pt}
    \begin{align*}
    \mathrm{KL}_{\tau+1} &= \log(|\Lambda_{\tau+1}\Lambda_{\tau+1}^\top + \Psi_{\tau+1}|) 
    \\
    &\qquad + \mathrm{tr}\big((\Lambda_{\tau+1}\Lambda_{\tau+1}^\top + \Psi_{\tau+1})^{-1} \SigmaFull \big)
    \end{align*}
    \vspace{-10pt}
    \IF{$(\mathrm{KL}_{\tau+1}/\mathrm{KL}_{\tau}< 1\!+\!\epsilon$)}
    \STATE $\, \mathbf{break}$   \COMMENT{\textcolor{MidnightBlue}{Early stopping}}
    \ENDIF
   \ENDFOR 
   \STATE \textbf{Output:} $\Sigma_{t+1}$ (represented implicitly by $\Lambda_{\tau+1}$,$\Psi_{\tau+1})$
\end{algorithmic}
\end{algorithm}

\section{The patch: a low-rank plus diagonal covariance matrix}
\label{sec:patch}

In this section, we consider how BaM can be adapted to a Gaussian variational family whose covariance matrices $\Sigma$ are parameterized as the sum of a low-rank matrix and a diagonal matrix: namely,
\begin{align}
    \Sigma =  \Lambda \Lambda^\top + \Psi.
    \label{eq:structured-cov}
\end{align}
In this parameterization, $\Lambda$ is a $D\!\times\!K$ matrix with rank $K\! \ll\! D$,
and $\Psi$ is a $D\!\times\! D$ diagonal matrix. For Gaussian distributions whose covariance matrices are of this form,
we can evaluate the log-density with $\mathcal{O}(D)$ computation using the Woodbury matrix identity. We can also generate samples efficiently as follows~\citep{miller2017variational, ong2018gaussian}:
\begin{align*}
    \epsilon \sim \mathcal{N}(0, I_D), \quad 
    \zeta \sim \mathcal{N}(0, I_K), \quad
    z = \mu + \Lambda \zeta + \sqrt{\Psi}\epsilon.
\end{align*}
Our goal is to derive an algorithm for score-based VI with this variational family whose computational cost and memory also scale linearly in $D$. To do so, we introduce a \emph{patch step}
that projects the covariance update in \Cref{eq:sigmaupdate}
into one that has the form of \Cref{eq:structured-cov}.

The new algorithm is shown in \Cref{alg:pBaM}. Intuitively, at each iteration,
we use $\Sigma_{t+\scriptscriptstyle\frac{1}{2}}$ to denote the intermediate 
update of the \textit{unconstrained} covariance matrix computed by \Cref{eq:sigmaupdate}, and then we apply a patch that projects $\Sigma_{t+\scriptscriptstyle\frac{1}{2}}$ to a low-rank plus diagonal matrix,
\begin{align}
\label{eq:patch}
\Sigma_{t+1} =
\texttt{Patch}\left[\SigmaFull\right]
\end{align}
such that the updated covariance matrix is still of the form in \Cref{eq:structured-cov}, i.e. $\Sigma_{t+1} = \Lambda_{t+1}\Lambda_{t+1}^\top + \Psi_{t+1}$.
{Note that we perform this patch without constructing the $\Sigma_{t+1}$ matrix explicitly, but only by estimating the parameters $\Lambda_{t+1}$ and~$\Psi_{t+1}$.}
We compute this projection via an Expectation-Maximization (EM) algorithm that minimizes the KL divergence, $\text{KL}(q_{t+\scriptscriptstyle\frac{1}{2}}||q_{t+1})$,
with respect to the parameters $(\Lambda_{t+1},\Psi_{t+1})$. Here,
$q_{t+\scriptscriptstyle\frac{1}{2}}$ and $q_{t+1}$ are the Gaussian distributions with shared mean $\mu_{t+1}$ and covariance matrices $\SigmaFull$ and $\Sigma_{t+1}$.

The next sections show why a patch is needed at each iteration to perform this projection, describe how the projection is performed via an EM algorithm, and discuss the overall scaling of this new algorithm.

\subsection{Why add a ``patch step''?}
\label{ssec:whypatch}

In this section, we investigate the update of \Cref{eq:sigmaupdate} in detail and show why a patch step is need to scale linearly in the number of dimensions, $D$.
Recall that when $B \ll D$,
the matrix $U$ in \Cref{eq:U} of rank at most $B\!+\!1$ because it is constructed from a sum of $B\!+\!1$ rank-one matrices.
Here we note the following: when $\Sigma_t$ is the sum of a diagonal plus low-rank matrix, as in \Cref{eq:structured-cov}, then the matrix $V$ in \Cref{eq:V} can also be written as the sum of a diagonal plus low-rank matrix. In particular, we can write
\begin{equation}
    \label{eq:V=structured}
    V = \Psi_t + RR^\top,
\end{equation}
where the matrix $R\!\in\!\reals^{D\times(B+1+K)}$ is constructed by concatenating the column vectors
\begin{equation}
\label{eq:R}
R = \left[\sqrt{\tfrac{\lambda_t}{B}}(z_1\!-\!\bar{z}),...,\sqrt{\tfrac{\lambda_t}{B}}(z_B\!-\!\bar{z}),\sqrt{\tfrac{\lambda_t}{1+\lambda_t}}
    (\mu_t\!-\!\overline{z}),\Lambda_t\right].
\end{equation}
The construction of the tall but thin matrix $R$ in \Cref{eq:R} is analogous to the construction of the matrix $Q$ in \Cref{eq:Q}.

With these definitions, we can now realize certain gains in efficiency when the number of dimensions, $D$, is  large. First,  note that 
the covariance update of
\Cref{eq:sigmaupdate} can be written explicitly in terms of $R$ and~$Q$ as
\begin{align}
   \SigmaFull
   &\!=\! \Psi_t \!+\! R R^\top \!\!
   \!-\! (\Psi_t \!+\! R R^\top)^\top\!\! Q M Q^\top
   \!\!
   (\Psi_t \!+\! R R^\top),
   \label{eq:sigmaupdate-lr}
\end{align}
where we have introduced $M$ 
as shorthand for the 
bracketed $(B+1)\times (B+1)$ matrix $\big[\tfrac{1}{2} I \!+\! \left(Q^\top V Q + \tfrac{1}{4} I\right)^\frac{1}{2}\big]^{-2}$ that appears in \Cref{eq:sigmaupdate}.
Hence the updated covariance matrix $\Sigma_{t+\scriptscriptstyle\frac{1}{2}}$ can be implicitly represented by the matrices
$\Psi_t$, $Q$, $R$, and $M$,
which are respectively of size $D$ (along the diagonal), $D\times(B\!+\!1)$, 
$D\times(K\!+\!B\!+\!1)$, and $(B\!+\!1)\times(B\!+\!1)$;
i.e., the memory cost is $\O(D)$.
Second, we note that none of the matrix products in \Cref{eq:sigmaupdate-lr} are between dense high-rank matrices.
Hence, we can also evaluate \Cref{eq:sigmaupdate-lr} with a computational cost that is $\O(D)$; this is shown more fully in \Cref{ssec:cost}.

\subsection{\texttt{Patch}:\ projection via an EM algorithm}
\label{ssec-em-algo}

To perform the patch in
\Cref{eq:patch},
we aim to minimize the KL divergence
$\text{KL}(q_{t+\scriptscriptstyle\frac{1}{2}}||q_{t+1})$, 
with respect to the parameters $(\Lambda,\Psi)$, where
\begin{align}
q_{t+\scriptscriptstyle\frac{1}{2}} &= \N(\mu_{t+1}, \SigmaFull),\\
q_{t+1} &= \N(\mu_{t+1}, \Sigma_{t+1}),
\end{align}
and $\Sigma_{t+1} =\Lambda \Lambda^\top \!\!+ \Psi$.
This KL divergence is given by
\begin{align}
\label{eq:kl:factors}
    & \text{KL}(q_{t+\scriptscriptstyle\frac{1}{2}}||q_{t+1})
    \\ \nonumber
&=
    \!\tfrac{1}{2}
    \!\left[
    \log\!\left(\tfrac{|\Lambda\Lambda^\top\!\! + \Psi|}{|\SigmaFull|}\!\right) \!-\! D \!+\! \trace\left\{(\Lambda\Lambda^\top \!\!+ \Psi)^{-1}\SigmaFull\right\}
    \right].
\end{align}
Note that the KL divergence is independent of the mean $\mu_{t+1}$, which is the same for both distributions. Hence, for simplicity, and without loss of generality, we assume that the mean is zero in the following discussion.

To minimize this objective, we adapt a \textit{factor analysis model}, as it allows us to minimize this KL divergence by defining and minimizing a simpler auxiliary function.
Indeed, the update we consider is the  infinite data limit of the EM algorithm for factor analysis \citep{rubin1982algorithms,ghahramani1996algorithm,saul2000maximum}.

Let  $\tilde q_{t+\scriptscriptstyle\frac{1}{2}}(z) = \N(z \given 0, \SigmaFull)$.
Consider the following factor analysis model:
\begin{align}
\label{eq:generative}
    \zeta \sim \N_K(0, I),
    \qquad
    z \given \zeta \sim \N_D(\Lambda \zeta, \Psi),
\end{align}
where $\zeta \in \reals^K$ is a hidden variable and $K \ll D$, and
the subscript on the Gaussian $\N({\cdot})$
denotes the dimension of the multivariate variable.
Then the marginal distribution $q_\theta(z)$, where $\theta~=~(\Lambda,\Psi)$, is Gaussian with the structured covariance matrix $\Sigma~=~\Lambda \Lambda^\top + \Psi$:
\begin{align}
    q_\theta(z) = \N(0, \Lambda \Lambda^\top + \Psi).
\end{align}
We estimate a factored representation for~$\SigmaFull$ by 
minimizing the
KL divergence between $\tilde q_{t+\scriptscriptstyle\frac{1}{2}}$ and $q_\theta$ with respect to $\theta$ (which is equivalent to minimizing \Cref{eq:kl:factors} with respect to $\Lambda$ and $\Psi$).

As is standard in the EM algorithm,
we minimize an auxiliary function that upper bounds the KL divergence.
This auxiliary function is minimized by simple updates that also
monotonically decrease the KL divergence.
The auxiliary function is given by
\begin{align}
\label{eq:auxiliaryf}
    \A(\theta_\tau, \theta) \!:=\! \E_{\tilde q_{t+\scriptscriptstyle\frac{1}{2}}}\!\!\left[
    -\!
    \!\int\! \log({q(\zeta, z \!\given\! \theta)})
     q(\zeta\!\given\! z, \theta_\tau)
     d\zeta\right]. 
\end{align}
We show in \Cref{ssec:auxfunction} how to minimize this function with respect to $\theta$, i.e.,
\begin{align}
\label{eq:update-specific}
    \theta_{\tau+1} \!=\!
    \arg\min_{\theta \in \Theta}
    \A(\theta_\tau, \theta),
\end{align}
and we also show that this update decreases $\KL(\tilde q_{t+\scriptscriptstyle\frac{1}{2}}|| q_\theta)$ at every iteration.

The EM algorithm takes the following form.
In the E-step, we compute the following statistics (expectations) of the conditional distribution of the hidden variable $\zeta$ given $z$:
\begin{align*}
    \E_q[\zeta]
    = \beta_\tau z,
    \qquad
    \E_q[\zeta \zeta^\top] =
    \beta_\tau z z^\top \beta_\tau^\top + I - \beta_\tau\Lambda_\tau,
\end{align*}
where
$\beta_\tau :=  \Lambda_\tau^\top(\Lambda_\tau\Lambda_\tau^\top + \Psi_\tau)^{-1}$ is a $K \times D$ matrix that
can be computed efficiently via the Woodbury matrix identity.
In the M-step, we optimize the objective in
\Cref{eq:auxiliaryf},
which becomes
\begin{align}
\label{eq:loss:fn}
\L 
&:=
    \tfrac{1}{2}
    \E_{\tilde q_{t+\scriptscriptstyle\frac{1}{2}}}\big[
    z^\top \Psi^{-1} z
    - 2z^\top \Psi^{-1} \Lambda \E_{q}[\zeta]
     \\
    &\qquad+
    \trace(\Lambda^\top \Psi^{-1} \Lambda\E_{q}[\zeta \zeta^\top])
    \big]
    \nonumber
    +\tfrac{\log|\Psi|}{2}
    + \text{const.}
\end{align}
The EM updates are derived by setting
$\nabla_\Lambda \L=0$ and
$\nabla_\Psi \L=0$.
In this way, we derive the updates
\begin{align}
\Lambda_{\tau+1}
&=  \SigmaFull \beta_\tau^\top
    (\beta_\tau \SigmaFull \beta_\tau^\top + I - \beta_\tau\Lambda_\tau)^{-1},
    \\
\Psi_{\tau+1}
    &=
    \text{diag}((I -  \Lambda_{\tau+1} \beta_\tau) \SigmaFull) .
\end{align}
These steps are summarized in \Cref{alg:patch},
and a complete derivation is provided in \Cref{ssec:EM:derivations}.

\begin{figure*}[t]
        \centering
         \includegraphics[width=0.74\linewidth]{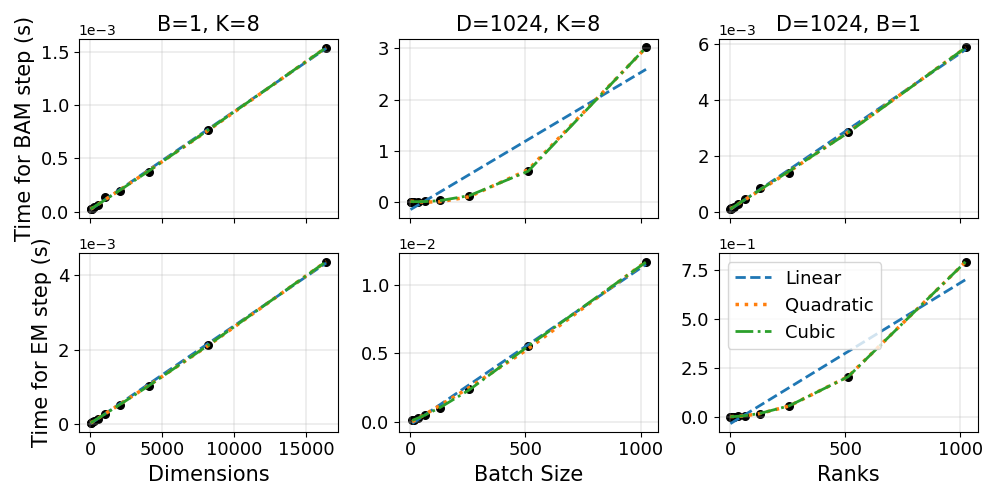}
    \caption{Scaling of pBaM algorithm with dimensions, batch size and rank. The top row shows the timing for BaM step (without evaluating the scores of the target) and the bottom row shows timing for a single EM step. All timings are in seconds. {We also show linear, quadratic and cubic fits to the data-points.}
    }
    \label{fig:scaling}
\end{figure*}

\subsection{Convergence of the EM updates}

To minimize the number of EM steps, \Cref{alg:patch} adds a momentum hyperparameter
$\eta$ to the EM updates (lines 7--8).
We also implement early stopping wherein we monitor the KL divergence (\Cref{eq:kl:factors}) for every update, and if the {relative} difference in KL over successive iterations is less than a threshold value, we terminate the patch step. {We fix this threshold value to $10^{-4}$ for all experiments in this paper.}
At every iteration of pBaM, we initialize $\Psi$ and $\Lambda$ for the patch step with their current values. As a result, the convergence is quite fast. Roughly speaking, we find that convergence requires hundreds of EM updates in the first couple iterations of pBaM (when the variational approximation is rapidly evolving), but fewer than 10 EM updates after only a few iterations of BaM. The average number of EM updates over 10,000 iterations was fewer than 5 in most of our experiments.

\subsection{Scaling of pBaM}

Before proceeding to numerical experiments, we study the scaling of our algorithm in terms of the dimensionality $D$, batch size $B$, and rank $K$.
Every iteration of our approach consists of two steps---a BaM update and the patch step, which is implemented as a series of EM updates.
In \Cref{ssec:cost}, we show that the BaM update scales as $\O(DB^2 + B^3 + KBD)$, while the EM update scales as $\O(DK^2 + K^3 + KBD)$.  
As desired, the scaling of both steps is linear in $D$.
We validate these scalings empirically in \Cref{fig:scaling}, where we show the time taken for the BaM update and a single EM step for increasing dimensionality, batch size, and rank, varying one of these factors while keeping the others fixed. For the BaM step, we use a dummy model in which it takes a negligible time to evaluate scores.

Note that while the EM updates scale similarly as each BaM step, the former are independent of the target distribution. On the other hand, for each BaM step, it is necessary to re-evaluate the scores (i.e., gradients) 
of the target distribution.
For any physical model of reasonable complexity, evaluating scores is likely to be far more expensive than a single iteration of EM.

\section{Related work}
\label{sec:related}

Structured covariances have been considered
in several variational inference algorithms.
In particular,
\citet{miller2017variational} and \citet{ong2018gaussian}
propose
using the Gaussian family with low rank plus diagonal
structured covariances as a variational family
in a BBVI algorithm
based on reparameterization of the ELBO objective.
\citet{tomczak2020efficient} study the use of
low-rank plus diagonal Gaussians for variational
inference in Bayesian neural networks.

\citet{bhatia2022statistical} study the problem
of fitting a low-rank plus diagonal Gaussian to
a target distribution; they show that when the target
itself is Gaussian, gradient-based optimization
of the (reverse) KL divergence results in a power iteration.
However, their approach restricts the diagonal to be constant.

Alternative families for structured VI have also
been proposed by many researchers
\citep{saul1995exploiting,hoffman2015structured,tan2018gaussian,ko2024provably}.
For example, \citet{tan2018gaussian} use sparse precision matrices to model conditional independence, and
\citet{ko2024provably} use covariance matrices with block diagonal structure.

\section{Experiments}
\label{sec:experiments}

The patched BaM algorithm (pBaM) is implemented in JAX \citep{jax2018github}, and publicly available on \href{https://github.com/modichirag/GSM-VI/}{Github}.
We compare pBaM against factorized and low rank plus diagonal ADVI (ADVI-D and ADVI-LR respectively) \citep{miller2017variational,ong2018gaussian},
both of which can scale to high dimensions.
For smaller dimensions, we will also compare against ADVI with full covariance (ADVI-F) and Batch-and-Match (BaM) with full covariance.
For higher dimensional experiments, we omit the full
covariance methods due to their 
large computational cost.
Unless otherwise mentioned, all experiments are done with batch size $B=32$.

\paragraph{pBaM Setup:} We  set the parameters of the patch step to default values of $\eta=1.2$ and tolerance $t=10^{-4}$. We experimented with different values and found the algorithm to be robust in the broad parameter ranges of $1<\eta<1.5$ and $t<10^{-3}$. We initialize the learning rate parameter as $\lambda_0=1$ and decay its value at each iteration according to the schedule $\lambda_t = \lambda_0/(1\!+\!t)$.

\paragraph{ADVI Implementation:} To implement ADVI we used automatic differentiation in JAX and 
the ADAM optimizer~\citep{kingma2014adam} to minimize the reverse KL divergence. In all experiments, we evaluate performance over a grid of learning rates and only show the best results. We scheduled the learning rate to decrease linearly over the course of iterations to a minimum value of $10^{-5}$. We also experimented with other forms of scheduling, e.g., cosine scheduling, but did not find significant differences in performance.

We implemented factorized and multivariate normal distributions in NumPyro.
We implemented a more efficient alternative for Gaussian distributions with low-rank plus diagonal covariance matrices. This implementation generates samples via the factor analysis model in \Cref{eq:generative} and evaluates the log-probability using the Woodbury matrix identity; the full covariance matrix is never stored in memory.

\subsection{Synthetic Gaussian targets}

\begin{figure*}[t]
    \centering
    \includegraphics[width=1.0\linewidth]{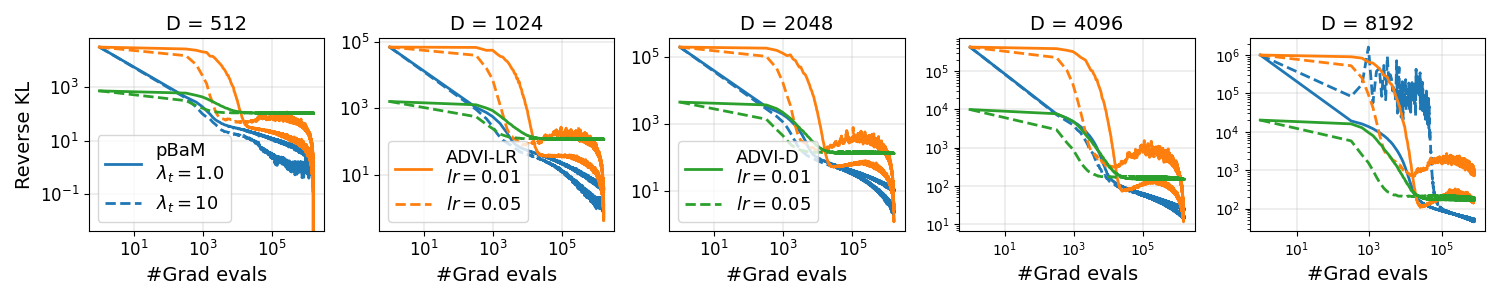}
    \caption{Performance with increasing dimensions for Gaussian with low-rank ($K=32$) plus diagonal covariance as target. We monitor reverse KL divergence for pBaM (blue) and ADVI-LR (orange) with $K=32$, and ADVI-D (green) for two different $\lambda_t$ and learning rate respectively (solid vs dashed).
    }
    \vspace{-10pt}
    \label{fig:scaling-dims}
\end{figure*}

We begin by considering Gaussian target distributions with a low-rank plus diagonal covariance:
\begin{align}
    p(z) &= \mathcal{N}(z \given \mu, \Psi + \Lambda\Lambda^\top), \nonumber \\
    \mu \sim \mathcal{N}(0, 1), \quad &\Psi \sim \mathcal{U}(0, 1),\quad
    \Lambda \sim \mathcal{N}(0, 1),
    \label{eq:gaussiantarget}
\end{align}
where $\mu,\Psi\! \in\! \reals^D$ and $\Lambda\! \in\!\reals^{D\times K}$.
Note that this distribution has strong off-diagonal components and the condition number $(\kappa)$ of this covariance matrix increases with increasing dimensions, e.g., $\kappa > 10^6$ for  $D>2048$ and $K=32$. We study the performance of pBaM on this target distribution as a function of the number of dimensions ($D$) and rank ($K$).

\subsubsection{Performance with increasing dimensions}

We begin by considering the setting where the variational distribution and the target distribution are in the same family. In this setting, we show that despite the patch step, the algorithm converges to match the target, even in high dimensions.

\Cref{fig:scaling-dims} shows the results on Gaussian targets with increasing numbers of dimensions but fixed rank $K\!=\!32$.
We show results for pBaM and ADVI-LR  that fit this target distribution with variational approximations of the same rank, $K=32$.
We monitor the reverse KL divergence as a function of gradient evaluations. Both algorithms are able to fit the target distribution, but pBaM converges much faster. We also show results for the best two learning rates for ADVI and two different values of $\lambda_t$ for pBaM.
While ADVI-LR converges faster initially with a large learning rate, it gets stuck closer to the solution even when the learning rate decays with a linear schedule. On the other hand, BaM performs well across different learning rates, though  though $\lambda_t=1$ is more stable than $\lambda_t=10$ for $D\!=\!8192$.

\subsubsection{Performance with increasing ranks}
\begin{figure*}[t]
    \centering
    \includegraphics[width=1.0\linewidth]{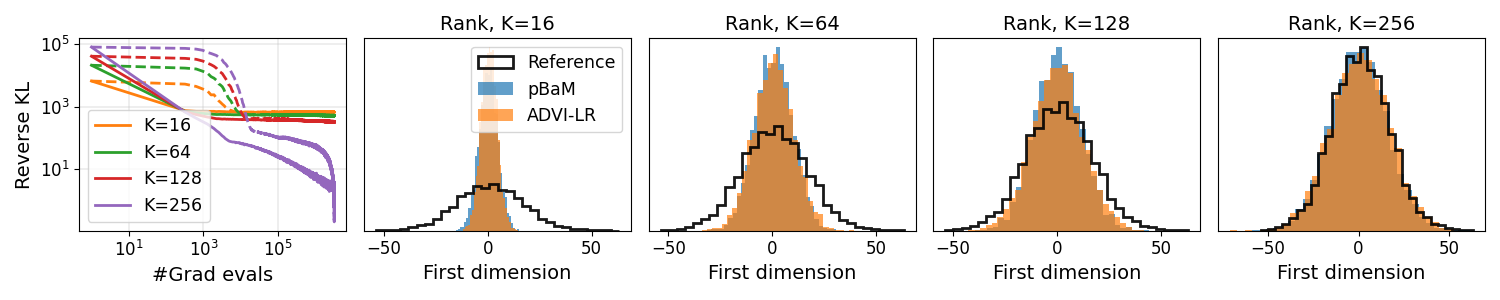}
    \caption{Performance with increasing ranks ($K$) for variational family. Target is 512 dimension Gaussian with low-rank ($K=256$) plus diagonal. We monitor reverse KL divergence for pBaM (solid) and ADVI-LR (dashed), and show marginal histograms for first dimension.
    }
     \vspace{-5pt}
    \label{fig:scaling-ranks}
\end{figure*}

Next we study the performance of pBaM with increasing ranks for the variational approximation. The results are shown in \Cref{fig:scaling-ranks}.
Here we have fixed the target to be Gaussian (\Cref{eq:gaussiantarget}) with $D=512$ and $K=256$, and we fit it with variational approximations of ranks $K=16, 64, 128$ and $256$. In the last setting, the variational approximation and the target again belong to the same family, and we expect an exact fit.

In the left panel of \Cref{fig:scaling-ranks}, we monitor the reverse KL divergence while in the other panels we show marginal histograms for the first dimension.
Again, we find that while both ADVI-LR and pBaM converge to the same quality of solution, pBaM converges much faster.
When the rank of the variational distribution is less than the target, the final approximation is more concentrated (smaller marginal variance). With increasing rank, the final approximation becomes a better fit, both in terms of minimizing the reverse KL and capturing the variance of the target distribution.

\subsection{Gaussian process inference}

\begin{figure}[t]
    \centering
    \includegraphics[width=0.95\linewidth]{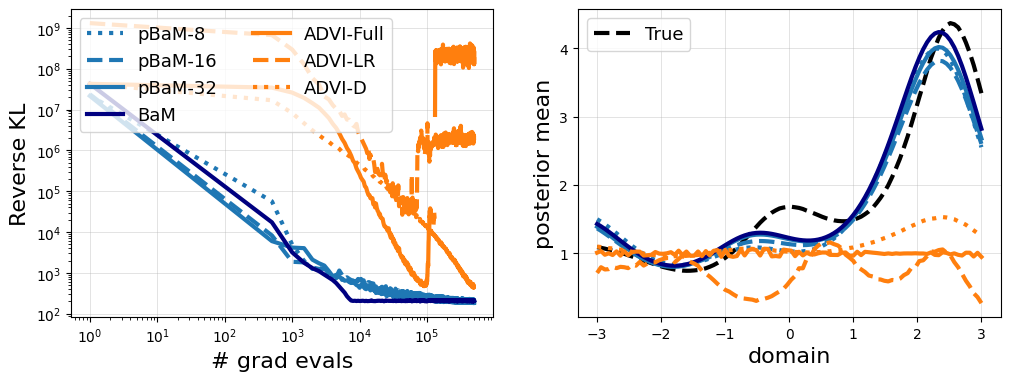}
    \caption{Synthetic GP-Poisson regression, $D=100$.
    We show the reverse KL (left)
    and the estimated rate using the posterior means (right).
    }
    \vspace{-5pt}
    \label{fig:poisson}
\end{figure}

\begin{figure}[t]
    \centering
    \includegraphics[width=\linewidth]{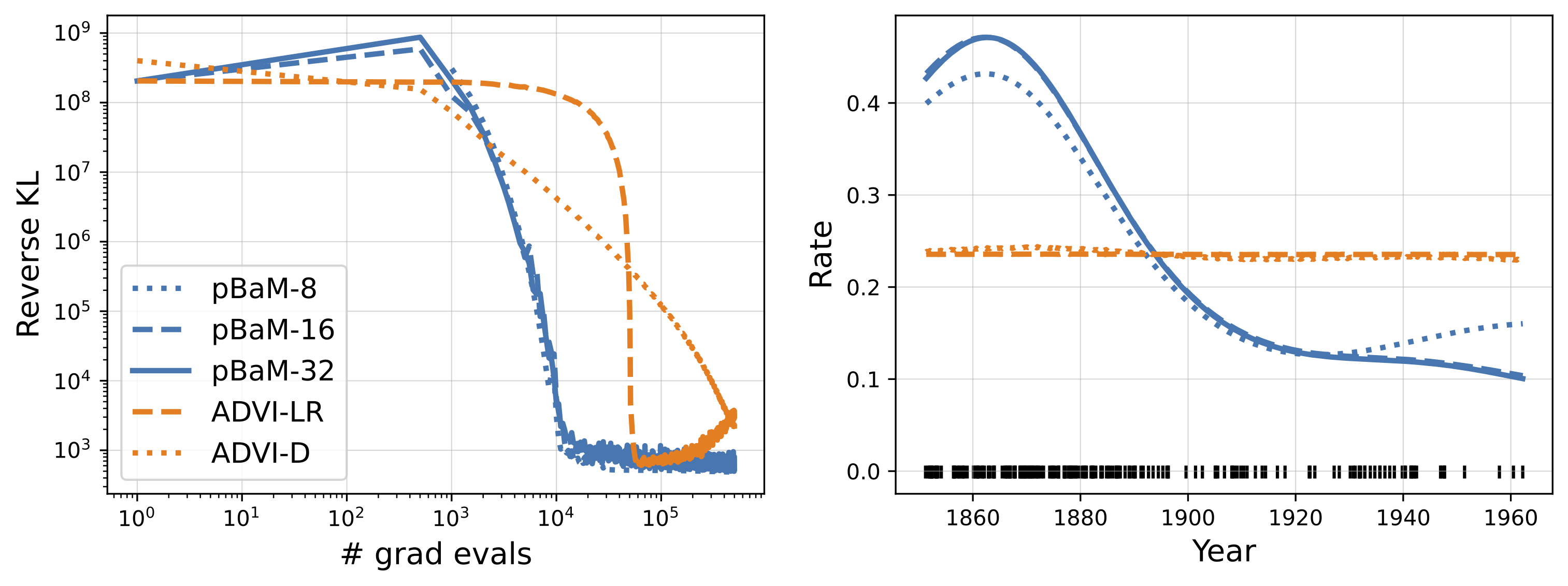}
    \caption{Log Gaussian Cox Process, $D=811$.
    We plot the reverse KL (left)
    and the estimated rate (right).
    }
     \vspace{-10pt}
    \label{fig:lgcp}
\end{figure}

Next we study pBAM for variational inference in Gaussian processes (GPs). Consider a generative model of the form
\begin{align}
    f \sim \text{GP}(m, \tilde\kappa),
    \quad y_{1:N} | f &\stackrel{iid}{\sim} P_f,
\end{align}
where
$m$ is a mean function, $\tilde\kappa$ is
a kernel function, and
$P_f$ is a distribution that characterizes the likelihood of a data point.
This structure describes many popular models
with intractable posteriors,
such as Gaussian process regression (with e.g., Poisson or Negative Binomial likelihoods), Gaussian process classification,
and
many spatial statistics models (e.g.,
the log-Gaussian Cox process).

The function $f$ is typically represented as a
vector of function evaluations
on a set of $D$ points,
and thus can be itself high-dimensional
(even if the data are low dimensional).
In all experiments, each method was 
run for a fixed number of iterations,
and so not all methods converged using
the same computational budget.

\paragraph{Poisson regression.}
We consider
a  Poisson regression problem in $D=100$ dimensions,
where
\begin{align*}
    p(y_1,\ldots,y_N; f) =
    \prod_{n=1}^N
    \text{Pois}(y_n; \lambda_n),
    \quad \lambda_n=\exp(f_n).
\end{align*}
Data were generated from this model
using a GP with zero-valued mean function and an RBF kernel
with unit length-scale.

In \Cref{fig:poisson},
we compare pBaM with ranks $K=8, 16,  32$ against the BaM algorithm and
ADVI (with full, rank 32, and diagonal covariances).
In the left plot, we report the
reverse KL vs the number of gradient evaluations.
Here we observe that the low-rank BaM
algorithms converge almost as quickly
as the full-rank BaM algorithm.
We found that even with extensive tuning
of the learning rate,
the ADVI algorithms tended to perform poorly
for this model; the full covariance ADVI algorithm diverges.
In the right plot, we show the true
rate function along with the inferred rate
functions (we report the posterior mean
from the VI algorithms).
For the BaM/pBaM methods, we found that increasing
the rank results in a better fit.

\paragraph{Log-Gaussian Cox process.}

Now we consider the log-Gaussian Cox process (LGCP)
\citep{moller1998log},
which is used to model point process data.
One approach to approximate
the log-Gaussian
Cox process is to bin the points
and then to consider a Poisson likelihood
for the counts in each bin.
Specifically,
we consider
$p(y_1,\ldots,y_N; f) =  \prod_{n=1}^N   \text{Pois}(y_n; \exp(f(x_n)+m)$,
where $m$ represents a mean offset.

We apply this model to a coal mining disasters data set \citep{carlin1992hierarchical},
which contains 191 events
of mining explosions in Britain between March 15, 1851 and March 22, 1962.
Following \citet{murray2010elliptical},
we
bin the points into $811$ bins,
and so the latent space is $D=811$ dimensional; in addition,
we use an RBF kernel with a length-scale
of $37$ and set the mean to $m=\log(191/811)$.

The left panel of \Cref{fig:lgcp}
shows the results of ADVI-LR, ADVI-D, and pBaM on ranks $8,16,32$.
Here we report the reverse KL divergence,
and we find that pBaM converges much
faster than the ADVI methods.
Notably, the varying ranks for pBaM
perform similarly.
The right panel of \Cref{fig:lgcp}
shows the estimated
rate functions
using the posterior mean of the variational
inference algorithms; the event data are
visualized in the black rug plot points.
  Comparing the estimated intensities from pBaM and the ADVI methods, we also see that the former are much closer to highly accurate estimates obtained from MCMC \citep[Figure~4]{nguyen2014automated}.

\begin{figure}[t]
    \vspace{-7pt}
    \centering
    \includegraphics[width=0.9\linewidth]{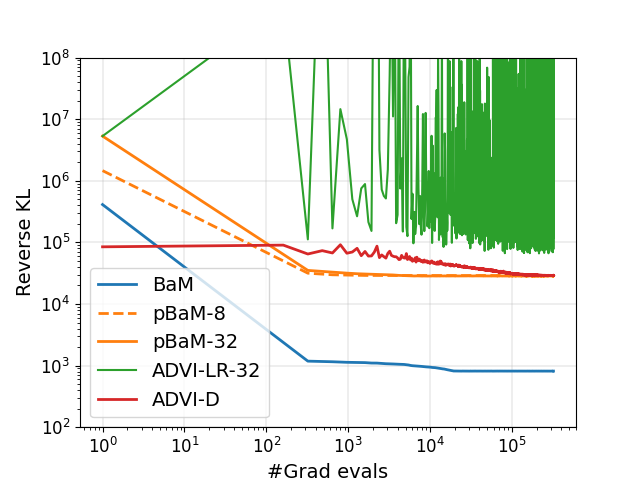}
    \includegraphics[width=0.9\linewidth]{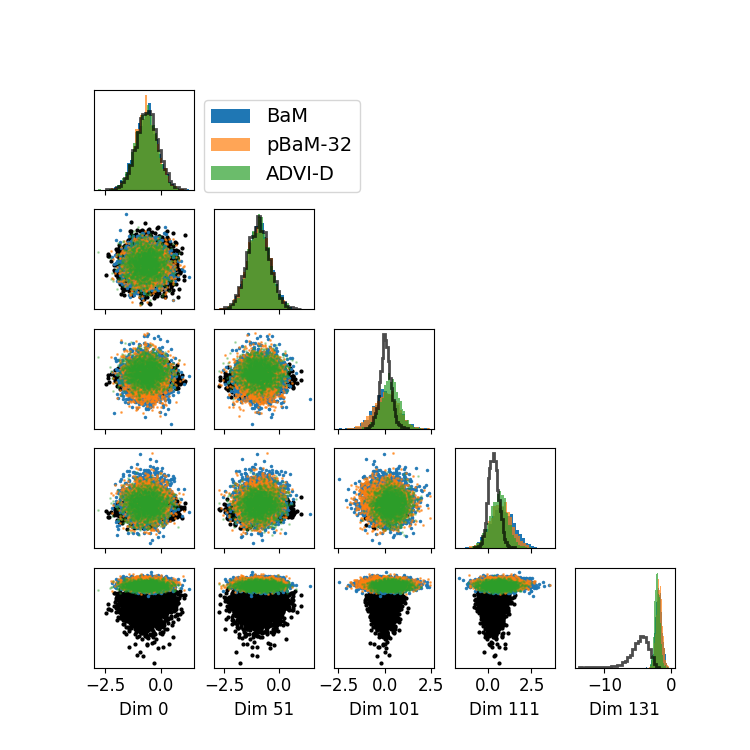}
    \caption{We show the reverse KL as well as 2-D marginals for five randomly selected dimensions for fitting IRT model with BaM, pBaM for ranks K=8 and 32, and ADVI-D. ADVI-LR and ADVI-F did not converge for this model.
    }
    \label{fig:irt}
\end{figure}

\subsection{IRT model}

Given $I$ students and $J$ questions, item response theory (IRT) models how students answer questions on a test depending on student ability ($\alpha_i$), question difficulty ($\beta_j$), and the discriminative power of the questions ($\theta_j$) \citep{gelman2006data}. The observations to be modeled 
are the binary values $\{y_{ij}\}$, where $y_{ij}$ indicates the correctness of the $i^\text{th}$ student's answer to question~$j$. Both the mean-centered ability of the student $\alpha_i$, and the difficulty of questions $\beta_j$ have hierarchical priors. These terms are combined into a variant of logistic regression with a multiplicative discrimination parameter.

 We show the results of variational inference for this model in \Cref{fig:irt}. Here $I=20$ and $J=100$, leading to dimensionality $D=143$. ADVI-LR and ADVI-F diverged on this model even with tuned learning rates. Full-rank BAM converges to a better reverse KL divergence than the low-rank pBaM or the factorized approximation of ADVI-D. However, the two-dimensional marginals for pBaM look closer to those of BAM than to those of the factorized approximation.

\section{Discussion and future work}
\label{sec:conclusion}

We have developed a
score-based BBVI algorithm for
high-dimensional problems where the covariance of the Gaussian variational family is represented
as a sum of a low-rank and diagonal (LR+D) component.
Our method extends the batch-and-match algorithm by adding a patch step that projects an unconstrained covariance matrix to one that is structured in this way.
The computational and storage costs of this
algorithm scale linearly in the dimension $D$.
Empirically, we find that our algorithm converges faster than ADVI with LR+D and is more stable on real-world examples.
We provide a Python implementation of pBaM at \href{https://github.com/modichirag/GSM-VI/}{https://github.com/modichirag/GSM-VI/}.

A number of future directions remain.
One is to study pBaM for other high dimensional
problems such as arise (for instance) in Bayesian neural networks.
Another is to successively increase the rank of variational approximation in pBaM 
after an initial low-rank fit.
Finally, it would also be of interest to consider other types of structured covariance matrices (e.g., exploiting sparsity) for scaling score-based VI to high dimensional distributions.

\section*{Acknowledgments}

We thank Robert Gower for helpful discussions.

\bibliographystyle{plainnat-mod}
\bibliography{main}

\begin{thebibliography}{31}
\providecommand{\natexlab}[1]{#1}
\providecommand{\url}[1]{\texttt{#1}}
\expandafter\ifx\csname urlstyle\endcsname\relax
  \providecommand{\doi}[1]{doi: #1}\else
  \providecommand{\doi}{doi: \begingroup \urlstyle{rm}\Url}\fi

\bibitem[Bhatia et~al.(2022)Bhatia, Kuang, Ma, and Wang]{bhatia2022statistical}
K.~Bhatia, N.~L. Kuang, Y.-A. Ma, and Y.~Wang.
\newblock Statistical and computational trade-offs in variational inference: A
  case study in inferential model selection.
\newblock \emph{arXiv eprint 2207.11208}, 2022.

\bibitem[Blei et~al.(2017)Blei, Kucukelbir, and McAuliffe]{blei2017vi}
D.~M. Blei, A.~Kucukelbir, and J.~D. McAuliffe.
\newblock Variational inference: A review for statisticians.
\newblock \emph{Journal of the American Statistical Association}, 112\penalty0
  (518):\penalty0 859--877, 2017.

\bibitem[Bradbury et~al.(2018)Bradbury, Frostig, Hawkins, Johnson, Leary,
  Maclaurin, Necula, Paszke, Vander{P}las, Wanderman-{M}ilne, and
  Zhang]{jax2018github}
J.~Bradbury, R.~Frostig, P.~Hawkins, M.~J. Johnson, C.~Leary, D.~Maclaurin,
  G.~Necula, A.~Paszke, J.~Vander{P}las, S.~Wanderman-{M}ilne, and Q.~Zhang.
\newblock {JAX}: composable transformations of {P}ython+{N}um{P}y programs,
  2018.

\bibitem[Cai et~al.(2024{\natexlab{a}})Cai, Modi, Margossian, Gower, Blei, and
  Saul]{cai2024b}
D.~Cai, C.~Modi, C.~Margossian, R.~Gower, D.~Blei, and L.~Saul.
\newblock Eigen{VI}: score-based variational inference with orthogonal function
  expansions.
\newblock In \emph{Advances in Neural Information Processing Systems},
  2024{\natexlab{a}}.

\bibitem[Cai et~al.(2024{\natexlab{b}})Cai, Modi, Pillaud-Vivien, Margossian,
  Gower, Blei, and Saul]{cai2024}
D.~Cai, C.~Modi, L.~Pillaud-Vivien, C.~Margossian, R.~Gower, D.~Blei, and
  L.~Saul.
\newblock Batch and match: black-box variational inference with a score-based
  divergence.
\newblock In \emph{International Conference on Machine Learning},
  2024{\natexlab{b}}.

\bibitem[Carlin et~al.(1992)Carlin, Gelfand, and Smith]{carlin1992hierarchical}
B.~P. Carlin, A.~E. Gelfand, and A.~F. Smith.
\newblock Hierarchical {B}ayesian analysis of changepoint problems.
\newblock \emph{Journal of the Royal Statistical Society: Series C (Applied
  Statistics)}, 41\penalty0 (2):\penalty0 389--405, 1992.

\bibitem[Gelman and Hill(2006)]{gelman2006data}
A.~Gelman and J.~Hill.
\newblock \emph{Data analysis using regression and multilevel/hierarchical
  models}.
\newblock Cambridge University Press, 2006.

\bibitem[Ghahramani and Hinton(1996)]{ghahramani1996algorithm}
Z.~Ghahramani and G.~E. Hinton.
\newblock The {EM} algorithm for mixtures of factor analyzers.
\newblock Technical report, Technical Report CRG-TR-96-1, University of
  Toronto, 1996.

\bibitem[Hoffman and Blei(2015)]{hoffman2015structured}
M.~D. Hoffman and D.~M. Blei.
\newblock Structured stochastic variational inference.
\newblock In \emph{Artificial Intelligence and Statistics}, pages 361--369,
  2015.

\bibitem[Jordan et~al.(1999)Jordan, Ghahramani, Jaakkola, and
  Saul]{jordan1999vi}
M.~I. Jordan, Z.~Ghahramani, T.~S. Jaakkola, and L.~K. Saul.
\newblock An introduction to variational methods for graphical models.
\newblock \emph{Machine Learning}, 37:\penalty0 183--233, 1999.

\bibitem[Kingma and Ba(2015)]{kingma2014adam}
D.~P. Kingma and J.~Ba.
\newblock Adam: A method for stochastic optimization.
\newblock In \emph{International Conference on Learning Representations}, 2015.

\bibitem[Kingma and Welling(2014)]{kingma2013auto}
D.~P. Kingma and M.~Welling.
\newblock Auto-encoding variational {B}ayes.
\newblock In \emph{International Conference on Learning Representations}, 2014.

\bibitem[Ko et~al.(2024)Ko, Kim, Kim, and Gardner]{ko2024provably}
J.~Ko, K.~Kim, W.~C. Kim, and J.~R. Gardner.
\newblock Provably scalable black-box variational inference with structured
  variational families.
\newblock In \emph{International Conference on Machine Learning}, 2024.

\bibitem[Kucukelbir et~al.(2017)Kucukelbir, Tran, Ranganath, Gelman, and
  Blei]{kucukelbir2017automatic}
A.~Kucukelbir, D.~Tran, R.~Ranganath, A.~Gelman, and D.~M. Blei.
\newblock Automatic differentiation variational inference.
\newblock \emph{Journal of Machine Learning Research}, 2017.

\bibitem[Miller et~al.(2017)Miller, Foti, and Adams]{miller2017variational}
A.~C. Miller, N.~J. Foti, and R.~P. Adams.
\newblock Variational boosting: Iteratively refining posterior approximations.
\newblock In \emph{International Conference on Machine Learning}, pages
  2420--2429. PMLR, 2017.

\bibitem[Modi et~al.(2023)Modi, Margossian, Yao, Gower, Blei, and
  Saul]{Modi2023}
C.~Modi, C.~Margossian, Y.~Yao, R.~Gower, D.~Blei, and L.~Saul.
\newblock Variational inference with {G}aussian score matching.
\newblock \emph{Advances in Neural Information Processing Systems}, 36, 2023.

\bibitem[M{\o}ller et~al.(1998)M{\o}ller, Syversveen, and
  Waagepetersen]{moller1998log}
J.~M{\o}ller, A.~R. Syversveen, and R.~P. Waagepetersen.
\newblock Log {G}aussian {C}ox processes.
\newblock \emph{Scandinavian Journal of Statistics}, 25\penalty0 (3):\penalty0
  451--482, 1998.

\bibitem[Murray et~al.(2010)Murray, Adams, and MacKay]{murray2010elliptical}
I.~Murray, R.~Adams, and D.~MacKay.
\newblock Elliptical slice sampling.
\newblock In \emph{Artificial Intelligence and Statistics}, volume~9, pages
  541--548, 2010.

\bibitem[Nguyen and Bonilla(2014)]{nguyen2014automated}
T.~V. Nguyen and E.~V. Bonilla.
\newblock Automated variational inference for {G}aussian process models.
\newblock \emph{Advances in Neural Information Processing Systems}, 27, 2014.

\bibitem[Ong et~al.(2018)Ong, Nott, and Smith]{ong2018gaussian}
V.~M.-H. Ong, D.~J. Nott, and M.~S. Smith.
\newblock Gaussian variational approximation with a factor covariance
  structure.
\newblock \emph{Journal of Computational and Graphical Statistics}, 27\penalty0
  (3):\penalty0 465--478, 2018.

\bibitem[Petersen and Pedersen(2008)]{petersen2008matrix}
K.~B. Petersen and M.~S. Pedersen.
\newblock The matrix cookbook.
\newblock \emph{Technical University of Denmark}, 7\penalty0 (15):\penalty0
  510, 2008.

\bibitem[Ranganath et~al.(2014)Ranganath, Gerrish, and
  Blei]{ranganath2014black}
R.~Ranganath, S.~Gerrish, and D.~Blei.
\newblock Black box variational inference.
\newblock In \emph{Artificial Intelligence and Statistics}, pages 814--822.
  PMLR, 2014.

\bibitem[Rubin and Thayer(1982)]{rubin1982algorithms}
D.~B. Rubin and D.~T. Thayer.
\newblock {EM} algorithms for {ML} factor analysis.
\newblock \emph{Psychometrika}, 47:\penalty0 69--76, 1982.

\bibitem[Saul and Jordan(1995)]{saul1995exploiting}
L.~Saul and M.~Jordan.
\newblock Exploiting tractable substructures in intractable networks.
\newblock \emph{Advances in Neural Information Processing Systems}, 8, 1995.

\bibitem[Saul and Rahim(2000)]{saul2000maximum}
L.~K. Saul and M.~G. Rahim.
\newblock Maximum likelihood and minimum classification error factor analysis
  for automatic speech recognition.
\newblock \emph{IEEE Transactions on Speech and Audio Processing}, 8\penalty0
  (2):\penalty0 115--125, 2000.

\bibitem[Tan and Nott(2018)]{tan2018gaussian}
L.~S. Tan and D.~J. Nott.
\newblock Gaussian variational approximation with sparse precision matrices.
\newblock \emph{Statistics and Computing}, 28:\penalty0 259--275, 2018.

\bibitem[Titsias and L{\'a}zaro-Gredilla(2014)]{titsias2014doubly}
M.~Titsias and M.~L{\'a}zaro-Gredilla.
\newblock Doubly stochastic variational {B}ayes for non-conjugate inference.
\newblock In \emph{International Conference on Machine Learning}. PMLR, 2014.

\bibitem[Tomczak et~al.(2020)Tomczak, Swaroop, and
  Turner]{tomczak2020efficient}
M.~Tomczak, S.~Swaroop, and R.~Turner.
\newblock Efficient low rank {G}aussian variational inference for neural
  networks.
\newblock \emph{Advances in Neural Information Processing Systems},
  33:\penalty0 4610--4622, 2020.

\bibitem[Wainwright et~al.(2008)Wainwright, Jordan,
  et~al.]{wainwright2008graphical}
M.~J. Wainwright, M.~I. Jordan, et~al.
\newblock Graphical models, exponential families, and variational inference.
\newblock \emph{Foundations and Trends{\textregistered} in Machine Learning},
  1\penalty0 (1--2):\penalty0 1--305, 2008.

\bibitem[Yang et~al.(2019)Yang, Martin, and Bondell]{yang2019variational}
Y.~Yang, R.~Martin, and H.~Bondell.
\newblock Variational approximations using {F}isher divergence.
\newblock \emph{arXiv preprint arXiv:1905.05284}, 2019.

\bibitem[Zhang et~al.(2018)Zhang, Shahbaba, and Zhao]{zhang2018variational}
C.~Zhang, B.~Shahbaba, and H.~Zhao.
\newblock Variational {H}amiltonian {M}onte {C}arlo via score matching.
\newblock \emph{Bayesian Analysis}, 13\penalty0 (2):\penalty0 485, 2018.

\end{thebibliography}

\appendix
\onecolumn
\section{The patch step:\ additional details}

\subsection{Derivation of the patch step update}
\label{ssec:EM:derivations}

In this section, we derive the infinite-data-limit
EM algorithm for factor analysis
in the context of pBaM.
In particular, we consider the model in
 \Cref{eq:generative}
 and the objective in
 \Cref{eq:update-specific}.

Recall that
$\tilde q_{t+\scriptscriptstyle\frac{1}{2}}(z) = \N(z \given 0, \SigmaFull)$,
where we assume we are given
$\SigmaFull$. As we discuss in
\Cref{ssec:whypatch}, we do not need to form this matrix explicitly.)

First note that the joint distribution under the model
in \Cref{eq:generative}
is
\begin{align}
    \begin{bmatrix}
        \zeta \\ z
    \end{bmatrix}
    \sim
    \N\left(
    \begin{bmatrix}
        0 \\ 0
    \end{bmatrix},
    \begin{bmatrix}
        I & \Lambda^\top \\
        \Lambda & \Lambda \Lambda^\top + \Psi
    \end{bmatrix}
    \right).
\end{align}

Defining
$\beta~:= ~\Lambda^\top(\Lambda \Lambda^\top + \Psi)^{-1}$,
the conditional distribution of $\zeta | z, \Psi,\Lambda$ is Gaussian
with mean and covariance
\begin{align}
\label{eq:condmean}
    \E_q[\zeta \given z]
    &= \Lambda^\top(\Lambda \Lambda^\top + \Psi)^{-1} z
    = \beta z
    \\
\label{eq:condcov}
    \text{Cov}_q[\zeta \given z]
    &=
    I - \Lambda^\top(\Lambda \Lambda^\top + \Psi)^{-1} \Lambda
    = I - \beta \Lambda.
\end{align}

In what follows, we denote the conditional expectation with respect to
$q(\zeta\given z, \Lambda_\tau,\Psi_\tau)$ as $\E_{q}[\cdot]$,
where $\tau$ represents the current EM iteration number.
In the E-step, we compute the following statistics of the conditional distribution (\Cref{eq:condmean} and \Cref{eq:condcov}) of the hidden variable $\zeta$ given $z$ and the current parameters
$\Lambda_\tau,\Psi_\tau$:
\begin{align}
\label{eq:estep:statistics}
    \E_q[\zeta] &=
    \beta_\tau z
    \\
    \E_q[\zeta \zeta^\top]
    &=
    \E_q[\zeta] \E_q[\zeta]^\top
    + \text{Cov}_q[\zeta]
    = \beta_\tau z z^\top \beta_\tau^\top + I - \beta_\tau \Lambda_\tau,
\end{align}
where
$\beta_\tau :=  \Lambda_\tau^\top(\Lambda_\tau\Lambda_\tau^\top + \Psi_\tau)^{-1}$.

In the M-step, we maximize the objective in
\Cref{eq:update-specific}.
First, we rewrite the objective
in terms of the optimization variables $(\Lambda,\Psi)$
and constants.
The log joint distribution is
\begin{align*}
    \log q(\zeta, z)
    &=
    -\tfrac{1}{2} (z-\Lambda\zeta)^\top \Psi^{-1} (z-\Lambda \zeta)
    -\tfrac{\log |\Psi|}{2} + \text{const.},
\end{align*}
and so the objective is
\begin{align}
\label{eq:loss:fn1}
\L 
&:=
\!-\E_{\tilde q_{t+\scriptscriptstyle\frac{1}{2}}}\!\left[
\E_{q}\!\left[
    \log\!\left({q(\zeta, z \given \Lambda,\Psi)}\!\right) \!
    \right] \right]
    + \text{const.},
    \\
    &=
    \tfrac{1}{2}
    \E_{\tilde q_{t+\scriptscriptstyle\frac{1}{2}}}\!\left[
    \E_{q}\!\left[
    z^\top \Psi^{-1} z
    -2z^\top \Psi^{-1} \Lambda \zeta
    +
    \zeta^\top\Lambda^\top \Psi^{-1} \Lambda \zeta
    \right] \right]
    +\tfrac{\log|\Psi|}{2}
    + \text{const.},
    \\
    &=
    \label{eq:objective:last}
    \tfrac{1}{2}
    \E_{\tilde q_{t+\scriptscriptstyle\frac{1}{2}}}\!\left[
    z^\top \Psi^{-1} z
    \!-\! 2z^\top \Psi^{-1} \Lambda \E_{q}[\zeta]
    \!+\!
    \trace(\Lambda^\top \Psi^{-1} \Lambda\E_{q}[\zeta \zeta^\top])
    \right]
    +\tfrac{\log|\Psi|}{2}
    + \text{const.},
\end{align}

Assuming we can exchange limits (and thus derivatives) and integration,
below, we compute the derivatives with respect
to the optimization parameters.
We use the following identities \citep{petersen2008matrix}:
\begin{align}
    \nabla_{\C} a^\top \C b &= a b^\top \\
    \nabla_{\C} \trace(\C A \C^\top) &= 2\C A \\
    \nabla_{\C} \log(|\C|) &= \C^{-1} \\
    \nabla_{\C} \trace\{\C B\} &= B^{\top},
\end{align}
where we assume above that $A=A^\top$.

\paragraph{Derivative w.r.t.\ $\Lambda$.}
Setting the derivative of
\Cref{eq:objective:last}
with respect to $\Lambda$
to 0, i.e.,
\begin{align}
    \nabla_\Lambda \L &=
    \E_{\tilde q_{t+\scriptscriptstyle\frac{1}{2}}}\left[\Psi^{-1} z \E_{q}[\zeta]^\top
    -
    \Psi^{-1} \Lambda \E[\zeta \zeta^\top]\right] = 0,
\end{align}
implies that
\begin{align}
    \Lambda^*
    &=
    \E_{\tilde q_{t+\scriptscriptstyle\frac{1}{2}}}[z \E_{q}[\zeta]^\top] \left(\E_{\tilde q_{t+\scriptscriptstyle\frac{1}{2}}}\E_q[\zeta \zeta^\top]\right)^{-1}
    \\
    &=
    \E_{\tilde q_{t+\scriptscriptstyle\frac{1}{2}}}[z z^\top] \beta_\tau^\top
    (\beta_\tau  \E_{\tilde q_{t+\scriptscriptstyle\frac{1}{2}}}[z z^\top] \beta_\tau^\top + I - \beta_\tau \Lambda_\tau)^{-1}
    \\
    &=
    \SigmaFull \beta_\tau^\top
    (\beta_\tau  \SigmaFull \beta_\tau^\top + I - \beta_\tau\Lambda_\tau)^{-1},
\end{align}
where in the second line we plugged
in the statistics from \Cref{eq:estep:statistics} (note that the statistics are evaluated with respect to the current
parameters $\Lambda_\tau, \Psi_\tau$),
and in the third line we used
$\E_{\tilde q_{t+\scriptscriptstyle\frac{1}{2}}}[zz^\top] = \SigmaFull$.

\paragraph{Derivative w.r.t.\ $\Psi$.}
Taking the derivative w.r.t.\ $\Psi^{-1}$
and setting $\nabla_{\Psi^{-1}}\L =0$, i.e.,
\begin{align}
    \tfrac{1}{2}
    \E_{\tilde q_{t+\scriptscriptstyle\frac{1}{2}}}\left[z z^\top \!\!-\! 2 \Lambda \E_q[\zeta] z^\top \!+\! \Lambda \E_q[\zeta \zeta^\top] \Lambda^\top\right] - \tfrac{1}{2} \Psi = 0
\end{align}
results in
\begin{align}
    \Psi^*
    &=
    \label{eq:deriv:psi}
    \E_{\tilde q_{t+\scriptscriptstyle\frac{1}{2}}}\left[z z^\top - 2 \Lambda \E_q[\zeta] z^\top+ \Lambda \E_q[\zeta \zeta^\top] \Lambda^\top\right]
    \\
    &=
    \E_{\tilde q_{t+\scriptscriptstyle\frac{1}{2}}}[z z^\top] -  \Lambda \E_{\tilde q_{t+\scriptscriptstyle\frac{1}{2}}}[\E_q[\zeta] z^\top]
    \\
    &=
    \SigmaFull -  \Lambda \beta_\tau \SigmaFull
    \\
    &=
    (I -  \Lambda \beta_\tau) \SigmaFull,
\end{align}
where the second line follows from substituting the $\Lambda^*$
update into the third term of
\Cref{eq:deriv:psi}.

The final update results from taking
the diagonal part of $\Psi^*$ above.
Note that this update depends on
the newly updated $\Lambda^*$ term in \Cref{eq:deriv:psi} (as opposed to the current $\Lambda_\tau$).

\subsection{The patch step updates monotonically improve KL divergence}
\label{ssec:auxfunction}
Suppose we have an auxiliary function
that has the following properties:
\begin{enumerate}[noitemsep,align=left]
    \item[\textbf{Property~1}:] $\A(\theta, \theta) = \KL(\tilde q_{t+\scriptscriptstyle\frac{1}{2}}, q_\theta)$
    \item[\textbf{Property~2}:] $\A(\theta, \theta') \geq \KL(\tilde q_{t+\scriptscriptstyle\frac{1}{2}}, q_{\theta'})$
    for all $\theta' \in \Theta$.
\end{enumerate}
Then the update
\begin{align}
\label{eq:update-general}
    \theta_{\tau+1} = \arg\min_{\theta \in \Theta}
    \A(\theta_\tau, \theta)
\end{align}
monotonically decreases $\KL(\tilde q_{t+\scriptscriptstyle\frac{1}{2}}; q)$
at each iteration.
The property that the KL divergence monotonically
decreases holds because
\begin{align}
    \KL(\tilde q_{t+\scriptscriptstyle\frac{1}{2}}, q_{\theta_\tau})
    \label{eq:prop1}
    &= \A(\theta_\tau, \theta_\tau)
    \\
    \label{eq:update}
    &\geq \A(\theta_\tau, \theta_{\tau+1})
    \\
     \label{eq:prop2}
    &\geq \KL(\tilde q_{t+\scriptscriptstyle\frac{1}{2}}, q_{\theta_{\tau+1}}),
\end{align}
where \Cref{eq:prop1} is  due to Property~1,
\Cref{eq:update} is by construction from the update,
and
\Cref{eq:prop2} is due to Property~2.

We now show that an auxiliary function satisfying Property 1 and 2 above exists.
In particular, consider
\begin{align}
\label{eq:auxfunction}
    \A(\theta,\theta') &=
    \E_{\tilde q_{t+\scriptscriptstyle\frac{1}{2}}}\!\left[\log \tilde q_{t+\scriptscriptstyle\frac{1}{2}}(z)
    \!-\!
    \!\int\! \log\!\left(\tfrac{q(\zeta, z \given \theta')}{q(\zeta\given z, \theta)}\!\right) \! q(\zeta\given z, \theta) d\zeta\right],
\end{align}
where $q(\zeta \given z, \theta) \propto q(\zeta) q(z\given \zeta, \theta)$ is the
conditional distribution of the hidden variable induced by
\Cref{eq:generative}.

If $\theta' = \theta$, then Property~1 is satisfied:
since $\log \frac{q(\zeta, z \given \theta)}{q(\zeta\given z, \theta)} = \log q_\theta(z)$ and $\int q(\zeta\given z) dz = 1$,  this results in $\A(\theta,\theta) = \KL(\tilde q_{t+\scriptscriptstyle\frac{1}{2}}, q_\theta)$.

Property~2 can be verified using an application of
Jensen's inequality to the second term of
\Cref{eq:auxfunction}:
\begin{align*}
    \A(\theta,\theta')
    &\geq
    \E_{\tilde q_{t+\scriptscriptstyle\frac{1}{2}}}\!\left[\log \tilde q_{t+\scriptscriptstyle\frac{1}{2}}(z)
    \!-\!
    \log\! \int\! \!\left(\tfrac{q(\zeta, z \given \theta')}{q(\zeta\given z, \theta)}\!\right) \! q(\zeta|z, \theta) d\zeta\right]
    \\
    &=
    \E_{\tilde q_{t+\scriptscriptstyle\frac{1}{2}}}\!\left[\log \tilde q_{t+\scriptscriptstyle\frac{1}{2}}(z)
    \!-\!
    \log\! \int\! \!{q(\zeta, z \given \theta')}d\zeta\right]
    \\
    &=
    \KL(\tilde q_{t+\scriptscriptstyle\frac{1}{2}}, q_{\theta'}).
\end{align*}

Thus, minimizing the function in
\Cref{eq:auxfunction} monotonically decreases the KL divergence.

\paragraph{Summary of objective and updates.}
Combining
\Cref{eq:update-general}
and
\Cref{eq:auxfunction}
results in the iteration
in
\Cref{eq:update-specific}: i.e.,
\begin{align}
    \theta_{\tau+1} \!=\! \arg\min_{\theta \in \Theta}
    \E_{\tilde q_{t+\scriptscriptstyle\frac{1}{2}}}\!\left[
    \!-\!\!
    \!\int\! \log({q(\zeta, z \given \theta)})
     q(\zeta\given z, \theta_\tau)
     d\zeta\right]
    \!\!+\! \text{const.},
\end{align}
where the
constant represents factors that are
independent of $\theta$.

\subsection{Computational cost of BaM update and the patch step}
\label{ssec:cost}

We first show that the 
computational cost of the 
BaM update with a structured
covariance is linear in the dimension $D$.
Next we discuss the computational cost
of the patch step, which in combination
with the BaM update, results in
a computational cost linear in $D$.

Recall from \Cref{eq:sigmaupdate-lr} that  the covariance can be written as
\begin{align*}
   \SigmaFull
   &= \Psi_t + R R^\top
   - (\Psi_t + R R^\top)^\top Q M Q^\top
   (\Psi_t \!+\! RR^\top),
\end{align*}
where $Q~\in~\reals^{D \times (B+1)}$
and $R~\in~\reals^{D\times(K+B+1)}$, and $\Psi_t$ is a $D$-dimensional diagonal matrix.

First consider $H:=(\Psi_t + RR^\top)^\top Q$,
where $H \in \reals^{D\times(B+1)}$;
this matrix can be constructed with 
$\mathcal{O}((K+B)BD)$ 
computation.

Then we store $\SigmaFull$ implicitly by storing the matrices $\Psi_t,\,R,H$ and $M$:
\begin{align}
   \SigmaFull
   &= \Psi_t + R R^\top 
   - H M H^\top,
  \label{eq:sigmaupdate-lrH}
\end{align}
Note that evaluating $R R^\top$ can be a $\mathcal{O}(D^2)$ operation, but since we store $R$, we do not need to explicitly evaluate it now, and we will be able to contract it with other elements in the patch update to maintain $\mathcal{O}(D)$ computation.

The matrix $M\in~\reals^{(B+1)\times(B+1)}$ defined in the update is:
\[M:=\big[\tfrac{1}{2} I_{B+1} \!+\! \left(Q^\top (\Psi_t + R R^\top) Q + \tfrac{1}{4} I_{B+1}\right)^\frac{1}{2}\big]^{-2} = \big[\tfrac{1}{2} I_{B+1} \!+\! \left(H^\top Q + \tfrac{1}{4} I_{B+1}\right)^\frac{1}{2}\big]^{-2}.\]
Here evaluating the matrix in the square-root takes $\mathcal{O}(DB^2)$ computation, and it takes $\O(B^3)$ to compute the square root and inverse.
Thus, overall, the total computational cost of the BaM update is  $\O(DB^2 + B^3 + KBD)$;
that is, it is linear in $D$.

Next, we turn to the computational cost of the patch step. Every iteration of patch step requires us to compute:
\begin{align*}
    \beta_\tau &=  \Lambda_\tau^\top
    \Psi_\tau^{-1} [I_K 
    \!-\! 
    \Lambda_\tau(I+\Lambda_\tau^\top \Psi_\tau^{-1} \Lambda_\tau)^{-1} \Lambda_\tau^\top \Psi_\tau^{-1}] 
    \quad \text{(via~Woodbury~inverse~identity)}
    \\
    \Lambda_{\tau+1} &= \SigmaFull \beta_\tau^\top
    (\beta_\tau  \SigmaFull \beta_\tau^\top + I_K - \beta_\tau\Lambda_\tau)^{-1}, \\
    \Psi_{\tau + 1} &= 
    \mathrm{diag}((I_D -  \Lambda_{\tau+1} \beta_\tau) \SigmaFull)
\end{align*}
Here $\beta_\tau \in \reals^{K \times D}$,
and
evaluating $\beta_\tau$ requires 
forming matrix products that take $\O(K^2D$) 
computation
and inverting a $K \times K$ matrix, 
which in total takes
$\mathcal{O}(K^2D + K^3)$ computation.

Using \Cref{eq:sigmaupdate-lrH}, $\SigmaFull \beta_\tau^\top$ can be evaluated in $\mathcal{O}(DK(K+B))$, and hence 
the $\Lambda_\tau$ update can be evaluated in $\mathcal{O}(DK(K+B) + K^3)$ computation,
with a similar complexity for the $\Psi$-update.

\section{Additional experiments}

In \Cref{tab:summary},
we provide an overview of the problems studied
in this paper.

\begin{table}[h]
    \centering
      \caption{Summary of problems
    }
    \begin{tabular}{ccc}
    \toprule
    \textbf{Problem} & \textbf{Target } & \textbf{$D$} \\
    \midrule
     Synthetic & Gaussian & 512  \\
     Synthetic & Gaussian & 1024  \\
     Synthetic & Gaussian & 2048  \\
     Synthetic & Gaussian & 4096  \\
     Synthetic & Gaussian & 8192  \\
     Poisson regression & Non-Gaussian &  100 \\
     Log Gaussian Cox process & Non-Gaussian  & 811 \\
     IRT model & Non-Gaussian & 143 \\
     \bottomrule
    \end{tabular}

    \label{tab:summary}
\end{table}

\subsection{Impact of hyperparameters on Gaussian example}
Here we show the impact of varying choices of parameters like scheduling, learning rate, and batch size for ADVI-LR and pBaM on the synthetic Gaussian example. 
We always work in the setting where the rank of the low-rank component of the target and the variational distribution is the same ($K=32$).

\begin{figure}
    \centering
    \includegraphics[width=0.95\linewidth]{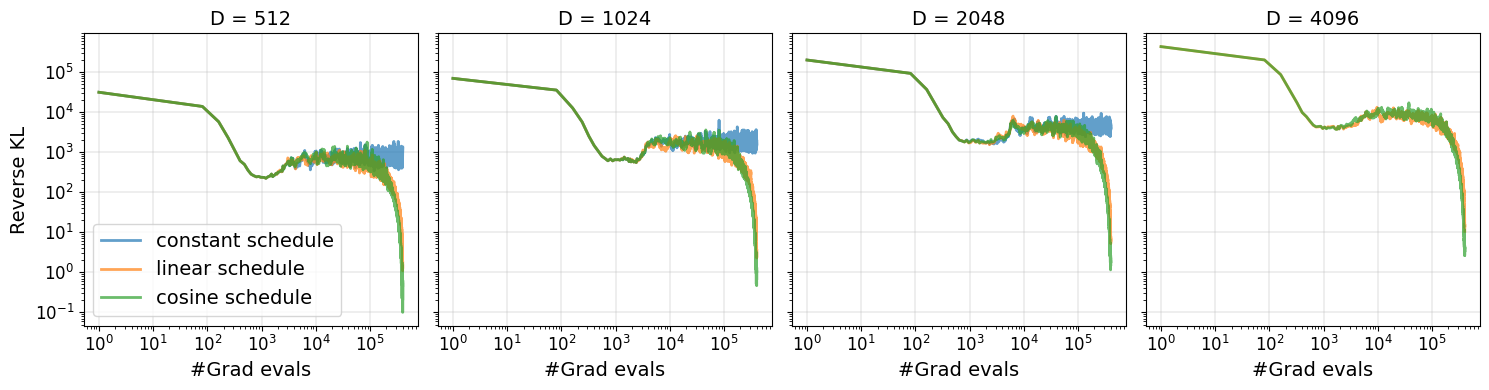}
    \caption{Impact of scheduling the learning rate for ADVI-LR.}
    \label{fig:advi-scheduling}
\end{figure} 
\Cref{fig:advi-scheduling} shows the impact of scheduling the learning rate. We consider no scheduling (constant learning rate), linear and cosine scheduling. We fix the inital and final learning rate to be $0.1$ and $10^{-5}$ respectively, and batch size $B=8$.
All three schedules converge at the same rate initially, however the constant learning rate asymptotes at a higher loss (KL divergence) once it reaches closer to the solution. Both linear and cosine scheduling give similar performance. Hence we use linear scheduling for all experiments.

\begin{figure}
    \centering
    \includegraphics[width=0.95\linewidth]{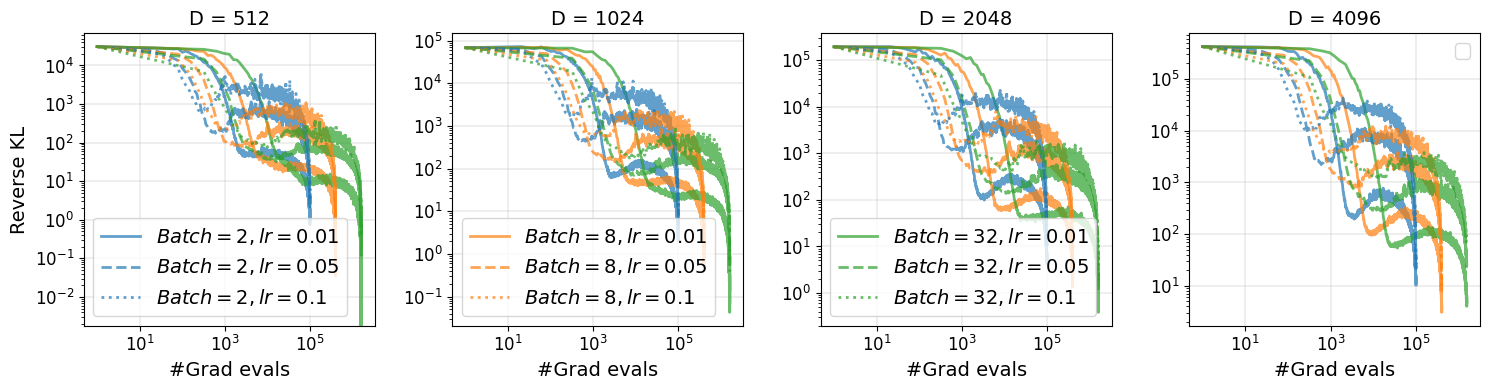}
    \caption{Impact of changing the batch size and the learning rate for ADVI-LR.}
    \label{fig:advi-batch-lr}
\end{figure}
Next in \Cref{fig:advi-batch-lr}, we study the impact of learning rate and batch size. Broadly, we find that the higher learning rate converges more quickly initially, but plateaus to a larger loss before decreasing sharply again. 
Smaller batch sizes show faster convergence initially, while they plateau at a lower KL divergence. Hence we use batch size $B=32$ for all the experiments. 
However, the conclusion of this figure along with \Cref{fig:advi-scheduling} suggests that the relationship with learning rate is more complicated.
Given that pBaM seems to outperform all these learning rates considered nevertheless, we show results for two best learning rates in the main text. 

\begin{figure}
    \centering
    \includegraphics[width=0.95\linewidth]{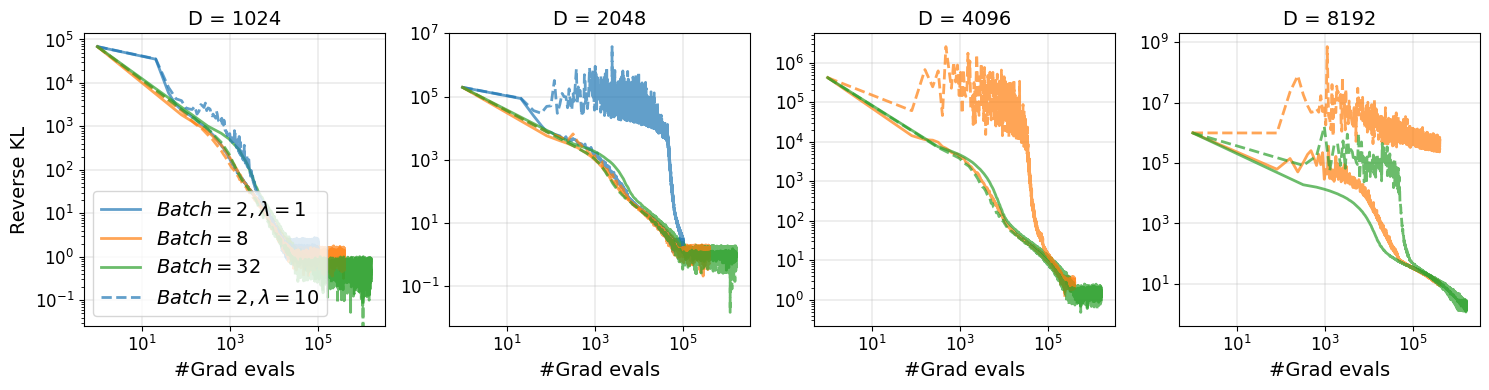}
    \caption{Impact of changing the batch size and the learning rate parameter $\lambda_t$ for pBaM.}
    \label{fig:bmp-batch}
\end{figure}
Finally we study the impact of batch size and learning rate parameter $\lambda_t$ for pBaM in \Cref{fig:bmp-batch}.
Overall, the algorithm seems to get increasingly more stable for larger batch size, and smaller $\lambda_t$ for a given batch size. However the convergence rate for any stable run seems to broadly be insensitive to these choices.

\subsection{Gaussian process inference}
\begin{figure}[h]
    \centering
    \includegraphics[width=0.32\linewidth]{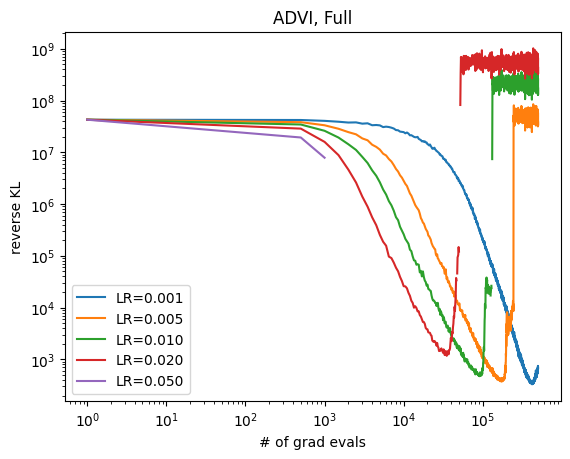}
    \includegraphics[width=0.32\linewidth]{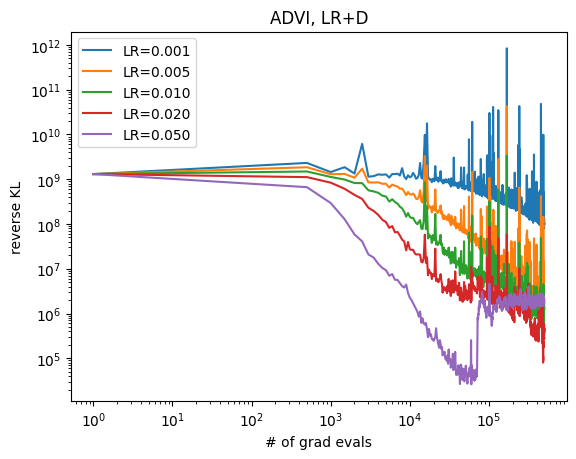}
    \includegraphics[width=0.32\linewidth]{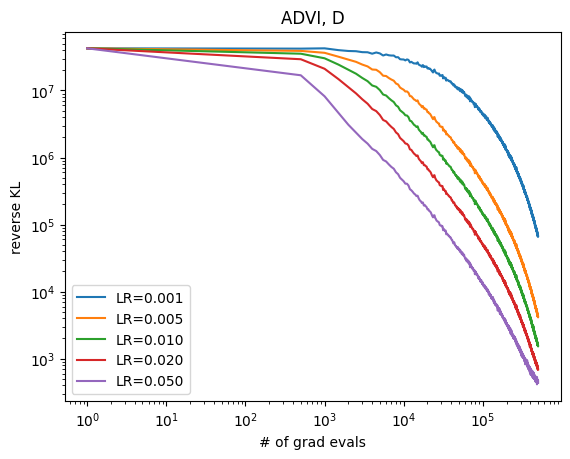}
    \caption{Poisson regression $D=100$ with varying learning rates for ADVI}
    \label{fig:poiss-reg-lr}
\end{figure}

\begin{figure}[h]
    \centering
    \includegraphics[width=0.95\linewidth]{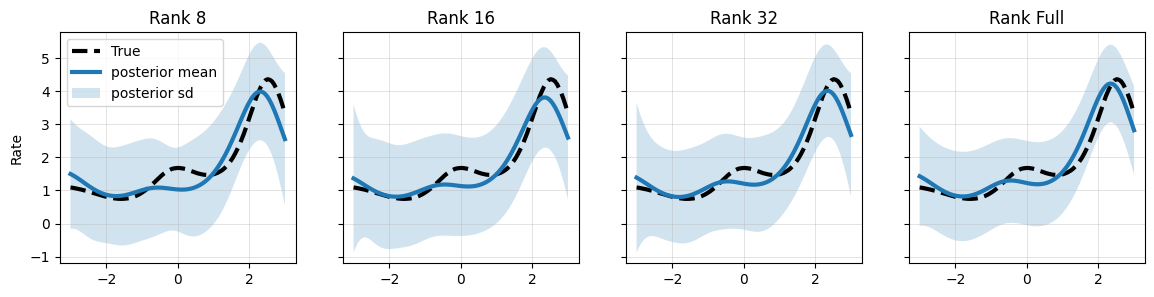}
    \caption{Poisson regression $D=100$. BaM/pBaM with varying ranks.}
    \label{fig:poiss-reg-rates}
\end{figure}

\subsubsection{Poisson regression}

In \Cref{sec:experiments}, we studied a $D=100$
Poisson regression problem.
First, we show in \Cref{fig:poiss-reg-lr}
the results of the different ADVI methods
on a range of learning rates: $0.001, 0.005, 0.01, 0.02, 0.05$.
For the final plots, the learning rate was set to 0.02.
For the full covariance family,
we also tried a linear schedule and found similar results.
We also found larger batch sizes to be more stable,
and so in all plots for this example, we set $B=50$.
In \Cref{fig:poiss-reg-rates}
we visualize the results of BaM/pBaM for varying ranks;
the estimated mean rates in the right panel of \Cref{fig:poisson},
but here we additionally show the uncertainties 
estimated from the posterior standard deviations.

\subsubsection{Log Gaussian Cox process}

\begin{figure}[h]
    \centering
    \includegraphics[width=0.33\linewidth]{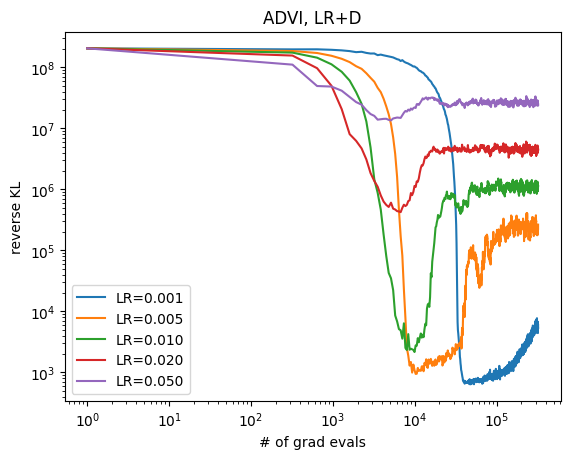}
    \includegraphics[width=0.33\linewidth]{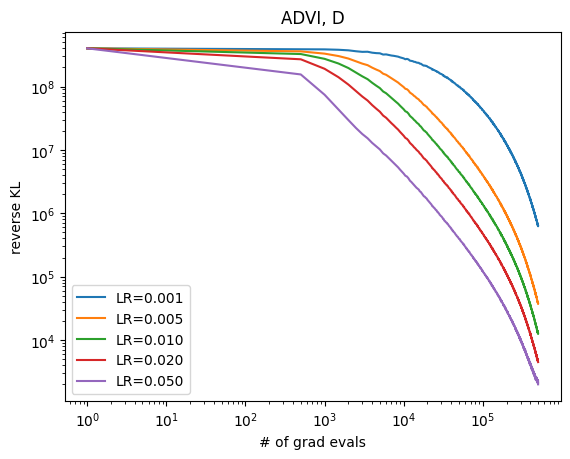}
    \caption{LGCP, $D=811$. ADVI with varying learning rate.}
    \label{fig:lgcp-lrs}
\end{figure}

\begin{figure}[H]
    \centering
    \includegraphics[width=0.95\linewidth]{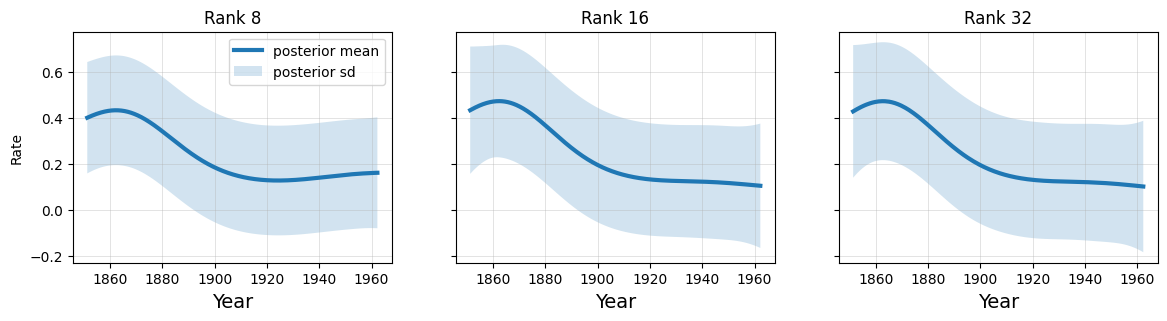}
    \caption{LGCP, $D=811$. BaM/pBaM with varying ranks.}
    \label{fig:lgcp-rates}
\end{figure}

In this example, the dimension was 811, and so we only
ran the variational inference methods with diagonal and low-rank plus diagonal families.
We show in  \Cref{fig:lgcp-lrs} ADVI for these families,
varying the learning rate on a grid.
In \Cref{fig:lgcp-rates}, we show the BaM/pBaM estimated mean rate 
and uncertainty computed using the variational posterior mean and standard deviation.
Overall, the differences in this set of
ranks were small.

\end{document}